\documentclass[runningheads]{llncs}
\usepackage{graphicx}
\usepackage{amsmath,amssymb}
\usepackage{graphicx}
\usepackage{multirow}
\usepackage{algorithm, algpseudocode}
\usepackage{booktabs} 
\usepackage{xcolor}
\usepackage{subcaption}
\usepackage{hyperref}

\usepackage{times}
\begin{document}
\title{Finding High-Value Training Data Subset through Differentiable Convex Programming}

%
\titlerunning{High-Value Training Data Selection}
%
\author{Soumi Das, Arshdeep Singh, Saptarshi Chatterjee \inst{1} \and
Suparna Bhattacharya\inst{2} \and
Sourangshu Bhattacharya\inst{1}
\institute{Indian Institute of Technology, Kharagpur \and
Hewlett Packard Enterprise, Bangalore}
}
\maketitle             
%
\begin{abstract}
Finding valuable training data points for deep neural networks has been a core research challenge with many applications. 
In recent years, various techniques for calculating the "value" of individual training datapoints have been proposed for explaining trained models.
However, the value of a training datapoint also depends on other selected training datapoints - a notion which is not explicitly captured by existing methods.
In this paper, we study the problem of selecting high-value subsets of training data.
The key idea is to design a learnable framework for online subset selection, 
which can be learned using mini-batches of training data, thus making our method scalable.
This results in a parameterised convex subset selection problem that is amenable to a differentiable convex programming paradigm, thus allowing us to learn the parameters of the selection model in an end-to-end training.
Using this framework, we design an online alternating minimization based algorithm for jointly learning the parameters of the selection model and ML model.
Extensive evaluation on a synthetic dataset, and three standard datasets, show that our algorithm finds consistently higher value subsets of training data, compared to the recent state of the art methods, sometimes $\sim 20\%$ higher value than existing methods.
The subsets are also useful in finding mislabelled training data. Our algorithm takes running time comparable to the existing valuation functions.

\keywords{Data valuation, Subset selection, Convex optimisation, Explainability}
\end{abstract}
\section{Introduction}
Estimation of "value" of a training datapoint from a Machine Learning model point of view, broadly called \textit{data valuation} \cite{ghorbani2019data,yoon2020data,pruthi2020estimating} , has become an important problem with many applications. One of the early applications included explaining and debugging training of deep neural models, where the influence of the training data points on the test loss is estimated \cite{koh2017understanding,pruthi2020estimating}. Another application involves estimating the expected marginal value of a training datapoint w.r.t. general value functions e.g. accuracy on a test set \cite{ghorbani2019data}, which can be used in buying and selling of data in markets. \cite{yoon2020data} lists a number of other applications, including detecting domain shift between training and test data, suggesting adaptive data collection strategy, etc. 

Most existing formulations for data valuation assume a supervised machine learning model, $y=f(x,\theta)$ with parameters $\theta$. For all the above mentioned applications, the key technical question is to determine the improvement in the value function (usually test set loss) given that a new set of datapoints are added to the training set.
The influence function based methods \cite{koh2017understanding,pruthi2020estimating} aim to estimate the influence of each individual datapoint, on the test loss of a trained model. While these techniques are computationally efficient, they ignore an important fact: \textit{the value of a training datapoint depends on the other  datapoints in the training dataset}. For a given training data point $x$, another datapoint $y$ may add to the value of $x$, possibly because it is very different from $x$. Alternately, $y$ may reduce the value of $x$ if it is very similar to $x$. 

Shapley value based methods \cite{ghorbani2019data} also estimate the value of individual datapoints using the expected marginal gain in the value function while adding the particular datapoint. Hence, this method considers the effect of other datapoints, but only in an aggregate sense. Also, this method is computationally expensive.
DVRL \cite{yoon2020data} is the closest to our approach. It learns a parameterised selection function which maximizes the reward of selecting training datapoints using Reinforcement Learning.
The reward function is the marginal improvement in value function after selection of a training datapoint over the previous running average of value function. Unfortunately, this reward leads to selection of many training datapoints which are  similar to each other, if their inclusion contributes to making the loss better than past average.
To the best of our knowledge, none of the existing data valuation techniques explicitly consider similarities between training points, when selecting high value subsets.

In this paper, we consider a set value function for assigning value to a subset of training datapoints. A proxy for the value function is learned though an embedding of datapoints, such that distance between the embedded datapoints represent the inverse value of replacing one of the datapoints with another.
In this framework, we propose the problem of \textit{high-value data subset selection}, which computes the optimal subset of a given size maximizing the value of the subset, or minimizing the total distance between selected points and other training datapoints in the embedding space.

In order to efficiently compute the high value data subsets, we address two challenges. Firstly, the embedding function of datapoints needs to be learned by optimizing the value of the selected subset, which itself is the output of an optimization problem. Hence, our formulation should be able to propagate the gradient of value-function back through the optimization problem. Secondly, for practicality, our formulation should be able to select the datapoints in an online manner from minibatches of data. 
We address both these challenges by designing a novel learned online subset selection formulation  for selecting subsets of training datapoints from minibatches. Our formulation is inspired by the online subset selection technique based on facility location objective \cite{elhamifar2017online}, but uses a learned distance function using a parameterised embedding of the training datapoints.
In order to learn the embedding parameters, our formulation is also compatible with the \textit{differentiable convex programming} paradigm \cite{agrawal2019differentiable}, which allows us to propagate the gradients of the value function (loss on test set) back to the embedding layer.
We propose an online alternating minimization based algorithm to jointly learn the embedding function for selection of optimal training data subset, and learn the optimal model parameters for optimizing the loss on the selected subset. Our algorithm scales linearly in training data size for a fixed number of epochs.

We benchmark our algorithm against state of the art techniques. e.g. DVRL \cite{yoon2020data}, Data Shapley \cite{ghorbani2019data}, etc. using extensive experiments on multiple standard real world datasets, as well as a synthetic dataset. We show that selection of high value subset of training data points using the proposed method consistently leads to better test set accuracy, for similar sized selected training dataset, with upto 20\% difference in test accuracy. Also,
removal of high value subset of training datapoints leads to a much more significant drop in test accuracy, compared to baselines.
We also show that the proposed method can detect higher fraction of incorrectly labelled data, for  all sizes of selected subsets. Moreover, correcting the detected mislabelled points leads to higher increase in test set accuracy compared to baseline.
We observe that DVRL \cite{yoon2020data}, which is the closest competitor, selects datapoints in a biased manner, possibly because there, it selects many similar high-performing datapoints at each stage.

To summarise, our key contributions are:
\begin{enumerate}
    \item We highlight the problem of high-value data subset selection, which selects a high-value data subset, rather than assigning valuation to individual datapoints.
    \item We formulate a novel learned online data subset selection technique which can be trained jointly with the ML model.
    \item We propose an online alternating minimization based algorithm which trains the joint model using mini-batches of training data, and scales linearly with training data.
    \item Results with extensive experimentation show that our method consistently outperforms recent baselines, sometimes with an accuracy improvement of $\sim 20\%$ for the same size subsets.
\end{enumerate}
\section{Related Work}
The exponential rise in quantity of data and depth of deep learning models have given rise to the question of scalability. Several subset selection methods have been designed using submodular \cite{buchbinder2014submodular} and convex approaches \cite{elhamifar2015dissimilarity}, to combat the problem. However, all these techniques only rely on some measure of similarity (pairwise, pointwise) \cite{elhamifar2017online}\cite{das2020multi} and do not really account for the explanation behind its choice in view of the end-task.

Besides scalability, the other question which arises is explainability. 
One set of methods surrounding explainability is prototype based approaches which finds the important examples aimed towards the optimisation of a given value function, or in other words, improvement of a given performance metric (e.g test set accuracy).
Pioneering studies on influence functions \cite{cook1982residuals} have been used to estimate the change in parameter $\theta$ , given an instance $x_i$ is absent and is used to obtain an estimate of $\theta_{-i} - \theta$. This aids in approximating the importance of the instance $x_i$ towards the performance of the learning model. \cite{koh2017understanding} have built upon the framework of influence functions and obtained an approximation of strongly convex loss functions. Following the works of \cite{koh2017understanding}, there have been several other works which try to measure the influence of training points on overall loss by tracking the training procedure \cite{hara2019data} while some try to measure the influence of training points on a single test point \cite{pruthi2020estimating} leveraging the training checkpoints. A recent work by \cite{wu2020deltagrad} develops an algorithm for rapid retraining of models using subset of data by reusing information cached during training phase.

Shapley value, an idea originated from game theory has been a driving factor in the field of economics. It was used as a feature ranking score for explaining black-box models \cite{lundberg2017unified}\cite{lundberg2018consistent}. However, \cite{ghorbani2019data} were the first to use shapley value to quantify data points where they introduce several properties of data valuation and use Monte Carlo based methods for approximation. There has also been a recent study in quantifying the importance of neurons of a black box model \cite{ghorbani2020neuron} based on shapley values. A follow-up work by \cite{ghorbani2020distributional} aims to speed up the computation of data shapley values. However, all the above methods do not have a distinctive selection mechanism for providing a representative set of points from incoming mini-batches which will be beneficial for explaining test set predictions.

Our work is in close proximity to the prototype based approaches where we intend to select the important training data points.  While \cite{ghorbani2020distributional} aims to work with training set distribution, our work is essentially inclined towards selection from training datasets, hence leading to choose \cite{ghorbani2019data} as one of our baselines. Among the methods around influence functions, we use the work of \cite{koh2017understanding} , \cite{pruthi2020estimating} as two of our baselines. A recent work by \cite{yoon2020data} attempts to adaptively learn data values using reinforcement signals, along with the predictor model. This work is the closest in terms of the problem objective we are intending to solve and hence we use it as one of our baselines. However, they intend to optimize the selection using reinforcement trick which requires a greater number of runs, while we use a learnable convex framework for the purpose of selection. We try to select the points using a learning selection framework which will be then used to optimize the subset through a differentiable value function. To the best of our knowledge, this is the first of its kind in using a differentiable convex programming based learning framework to optimize the selection mechanism in order to improve on a given performance metric.
\section{High-Value Data Subset Selection}
\newcommand{\cD}{\mathcal{D}}
\newcommand{\cX}{\mathcal{X}}
\newcommand{\cY}{\mathcal{Y}}
\newcommand{\cS}{\mathcal{S}}
\newcommand{\cL}{\mathcal{L}}
\newcommand{\cV}{\mathcal{V}}
\newcommand{\cO}{\mathcal{O}}
\newcommand{\psistar}{\Psi^*}
\newcommand{\thetastar}{\theta^*}
\newcommand{\argmin}{\mbox{arg}\min}
\newcommand{\psihat}{\hat{\Psi}}

In this section, we formally define the problem of high-value data subset selection and propose a learned approximate algorithm for the same.

\subsection{Motivation and Problem Formulation}

Data valuation for deep learning is an important tool for generating explanations of deep learning models in terms of training data.
Let $\cD = \{ (x_i,y_i) | i=1,\dots,n \}$ be the training dataset with features $x_i\in \cX$, and labels $y_i\in \cY$, and $\cD^t = \{ (x^t_i,y^t_i) | i=1,\dots,m \}$ be a test dataset, which is used for valuation of the training data in $\cD $. Also, let $s\in\cS$ denote a subset of of the training data $\cD$ ($\cS$ is the power set of $\cD$). Let $f(\theta)$ denote a parameterized family of learnable functions (typically implemented with neural networks), parameterized by a set of parameters $\theta$. Given an average loss function $\cL(f(\theta),s)$, defined as $\cL(f(\theta),s) = \frac{1}{|s|} \sum_{(x_i, y_i)\in s} L(y_i,f(x_i; \theta))$, we define the optimal parameters $\thetastar(s)$ for a training data subset $s$ as:
\begin{equation}
\thetastar(s) = \argmin_{\theta} \cL(f(\theta),s)
\label{eq:trloss1}
\end{equation}
Given an optimal parameter $\thetastar(s)$ and the test dataset $\cD^t$, we define the value function $v(s)$, which provides a valuation of the subset $s$ as:
\begin{equation}
v(s) = v(\thetastar(s),\cD^t) = \cL(f(\thetastar(s)),\cD^t) \label{eq:defvalue}
\end{equation}
The \textit{high-value subset selection problem} can be defined as:
\begin{equation}
\max_{s\in \cS} v(s) \ \mbox{ sub. to } |s|\leq \gamma n
\end{equation}
where $\gamma$ is the fraction of retained datapoints in the selected subset. 

This problem is NP-Hard for a general value function $v(s)$. Next, we describe our broad framework to approximate the above problem with a two step approach: 
\begin{enumerate}
\item Learn an embedding function $h(x,\phi)$ of the data point $x$, such that the distance between datapoints $x_i$ and $x_j$, $d_{ij}=d(x_i,x_j)=D(h(x_i,\phi), h(x_j,\phi))$ represents their inverse-ability to replace each other for a given learning task. Here, $D$ is a fixed distance function.
\item Select the set of most important datapoints $s^*$ by minimizing the sum of distance between each datapoint and its nearest selected neighbor, and update the parameters $\phi$ of the embedding function such that $v(s*)$ is maximized.
\end{enumerate}
The objective function for the selection problem in second step can be written as:
\begin{equation}
s^* = \min_{s\in\cS} \sum_{(x,y)\in\cD} \min_{(x',y')\in s} d(x,x')
\end{equation}
This objective function corresponds to the facility location problem \cite{elhamifar2017online}, which can be relaxed to the following convex linear programming problem:
\begin{align}
\min_{z_{ij}\in[0.1]} & \sum_{i,j=1}^n z_{ij} d(x_i,x_j)  \label{eq:opt1} \\
\mbox{sub. to} & \sum_{j=1}^n z_{ij} = 1, \ \forall i=\{1,\dots,n\} \nonumber
\end{align}
Here $z_{ij}=1$ indicate that the datapoint $j$ is the nearest selected point or ``representative'' for data point $i$. Hence, the objective function sums the total distance of each point from its representative point, and the constraint enforces that every datapoint $i$ should have exactly one representative point. A point $j$ is selected if it is representative to at least one point, i.e. $\sum_{i=1}^n z_{ij} \geq 1$. Let $u_j$ be the indicator variable for selection of datapoint $j$. For robustness to numerical errors, we calculate $u_j$ as:
\begin{equation}
u_j = \frac{1}{\xi} \max\{ \sum_{i=1}^n z_{ij}, \xi \} \label{eq:defu}
\end{equation}
where $0\leq \xi\leq 1$ is a constant.

There are two main challenges in utilizing this formulation for the problem of optimizing the value function. 
Firstly, the optimal parameters $\thetastar(s)$ depend on the variables $z_{ij}$ through variables $u_j$, which are output of an optimization problem. Hence, optimizing $v(s^*)$ w.r.t parameters ($\phi$) of the embedding function  requires differentiating through the optimization problem. 
Secondly, for most practical applications calculating the subset of training data, $s$, by optimizing the above optimization problem is too expensive for an iterative optimization algorithm which optimizes $\phi$. Moreover, optimizing the parameters ($\theta$) of the neural network model $f(\theta)$, typically involves an online stochastic optimization algorithm (e.g. SGD) which operates with mini-batches of training data points. In the next section, we describe the proposed technique for joint online training of both the learning model parameters $\theta$ and embedding model parameters $\phi$.

\subsection{Joint Online Training of Subset Selection and Learning Model}


Our problem of selection of high-value subsets from training data can be described as a joint optimization of value function for selection of best subset, and optimization of loss function on the selected subset computing the best model.
Given the training and test datasets $\cD$ and $\cD^t$ respectively, our overall value function parameterised by the 
embedding model parameters $\phi$ for selection was defined as (equation \ref{eq:defvalue}):
\begin{equation}
v(s(\phi)) = v(\thetastar(s(\phi)),\cD^t) = \frac{1}{|\cD^t|} \sum_{(x,y)\in\cD^t} L(y,f(\thetastar(s(\phi)),x)) \label{eq:defv2}
\end{equation}
where  $\thetastar(s(\phi)) = \argmin_{\theta} \cL(f(\theta),s(\phi)) $ (from equation \ref{eq:trloss1}). Representing $s(\phi)$ in terms of the selection variables $u_j$ (Eqn \ref{eq:defu}), we can write the model loss on selected training set using embedding model parameter $\phi$ as:
\begin{equation}
\cL(\theta;\phi,\cD) = \frac{1}{\gamma |\cD|} \sum_{(x_j,y_j)\in\cD} u_j(\phi) L(y_j,f(x_j;\theta)) \label{eq:defloss}
\end{equation}
We note that the above objective functions $\cL(\theta)$ and $v(s(\phi))$ are coupled in the variables $\phi$ and $\theta$. On one hand, the loss on the selected training set $\cL(\theta;\phi,\cD)$ depends on the current embedding model parameters $\phi$. On the other hand, intuitively, the value $v(s(\phi))$ of a selected set of datapoints $s(\phi)$ should also depend on the current model parameter value $\theta'$, which maybe be updated using loss from the selected set of datapoints $s(\phi)$. Hence, we define the cumulative  value function $\cV(\phi;\theta',\cD,\cD^t)$ as:
\begin{equation}
\cV(\phi;\theta',\cD,\cD^t) = v(\hat{\theta},\cD^t),\ \mbox{where}\ \hat{\theta} = \theta' - \alpha \nabla_{\theta} \cL(\theta;\phi,\cD) \label{eq:defCV}
\end{equation}
Here, $\hat{\theta}$ is the one step updated model parameter using the selected examples from dataset $\cD$ using parameter $\phi$, and $\alpha$ is the stepsize for the update. We combine the two objectives into a joint objective function, $J(\theta,\phi;\cD,\cD^t)$. The combined optimization problem becomes:
\begin{equation}
\thetastar, \phi^* = \argmin_{\theta,\phi} J(\theta,\phi;\cD,\cD^t) = \argmin_{\theta,\phi} ( \cV(\phi;\theta',\cD,\cD^t) + \cL(\theta;\phi,\cD)) \label{eq:jointdef}
\end{equation}
As discussed above, since the training dataset $\cD$ is normally much bigger in size, we design our algorithm to optimize the above objective function in an online manner. Hence the training dataset $\cD$ is split into $k$ equal-sized minibatches $\cD = \{ D_1,\dots,D_k\}$. The above joint objective function can be decomposed as:
\begin{equation}
J(\theta,\phi;\cD,\cD^t) = \frac{1}{k} \sum_{i=1}^k \{ \cL(\theta;\phi,D_i) + \cV(\phi;\theta',D_i,\cD^t) \}
\end{equation}
We solve the above problem using an online alternating minimization method similar to the one described in \cite{choromanska2019beyond}. Our algorithm updates $\theta$ and $\phi$ as:
\begin{eqnarray*}
 & \mbox{Initialize}\ \theta^1\ \mbox{and}&\ \phi^1 \ \mbox{randomly}\\
 &\mbox{for}\ t = 1,\dots , T & : \\
1: & & \theta^{t+1} = \theta^t - \alpha \frac{1}{k} \nabla_{\theta} \cL(\theta; \phi^t, D_{i(t)}) \\
2: & & \phi^{t+1} = \phi^t - \beta \frac{1}{k} \nabla_{\phi} \cV(\phi;\theta^t,D_{i(t)},\cD^t)
\end{eqnarray*}
Here $i(t)$ is the index of the minibatch chosen at update number $t$. Step 1 is the standard SGD update minimizing the loss over a subset of the minibatch $D_{i(t)}$ chosen using embedding parameters $\phi^t$. Hence, it can be implemented using standard deep learning platforms. Implementation of step 2 poses two challenges: (1) computation of $ \nabla_{\phi} \cV(\phi;\theta^t,D_{i(t)},\cD^t)$ requires differentiation through an optimization problem, since $u_j(\phi)$ is the result of an optimization. (2) the datapoints in $D_{i(t)}$ must also be compared with datapoints in a previous minibatch for selecting the best representative points. Hence we need an online version of the optimization problem defined in equation \ref{eq:opt1}. We describe solutions to these challenges in the next section.

\subsection{Trainable Convex Online Subset Selection Layer}
\label{sec:trainlayer}


In order to implement the algorithm outlined in the previous section, we have to compute  $\nabla_{\phi} \cV(\phi;\theta^t,D_{i},\cD^t)$, which can be done by combining equations \ref{eq:defv2},\ref{eq:defloss} and \ref{eq:defCV} as:
\begin{align}
\nabla_{\phi} \cV(\phi;\theta^t,D_{i},\cD^t) = & -\frac{1}{|\cD^t|} \sum_{(x_j,y_j)\in \cD^t} \nabla_{\hat{\theta}} l(y_j,f(\hat{\theta},x_j)) \cdot \label{eq:CVgrad}\\
 &\frac{1}{\gamma | D_i | } \sum_{(x_j,y_j)\in D_i} ( \nabla_{\phi} u_j(\phi) \nabla_{\theta} L(y_j,f(x_j,\theta)) \nonumber
\end{align}
The key challenge here is computation of $\nabla_{\phi} u_j(\phi)$, since $u_j(\phi)$ is the output of an optimization problem. Also, the optimization problem in equation \ref{eq:opt1} selects from all datapoints in $\cD$, whereas efficient loss minimization demands that we only select from the current minibatch $D_{i(t)}$. We use the online subset selection formulation, proposed in \cite{elhamifar2017online}. Let $\cO(t)$ denote the set of old datapoints at time $t$ which have already been selected, and let $D_{i(t)}$ denote the new set from which points will be selected. We use two sets of indicator variables: $z_{ij}^o=1$ indicating that old datapoint $j$ is a representative of new datapoint $i$, and $z_{ij}^n=1$ indicating that new datapoint $j$ is a representative of new datapoint $i$. Under this definition the optimization problem can written as:
\begin{align}
\min_{z^o_{ij},z^n_{ij}\in[0,1]} & \sum_{x_i\in D_{i(t)},x_j\in \cO(t)} z^o_{ij} d(x_i,x_j) + \sum_{x_i\in D_{i(t)},x_j\in D_{i(t)}} z^n_{ij} d(x_i,x_j)  \label{eq:opt2} \\
\mbox{sub. to} & \sum_{x_j\in\cO(t)}  z^o_{ij} + \sum_{x_j\in D_{i(t)}} z^n_{ij} = 1, \ \forall i=\{1,\dots,n\} \nonumber \\
 & \sum_{x_j\in D_{i(t)}} u_{j} \leq \gamma |D_{i(t)}| \nonumber
\end{align}
where $u_j = \frac{1}{\xi} \max\{ \sum_{x_i\in D_{i(t)}} z^n_{ij}, \xi \}$.
The first constraint enforces that all new points must have a representative, and the second constraint enforces that the total number of representatives from the new set is less than $\gamma |D_{i(t)}| $. The objective function minimizes the total distance of between all points and their representatives. As previously, $d(x_i,x_j)=D(h(x_i,\phi),h(x_j,\phi))$, where $h(x,\phi)$ is the embedding of the point $x$ using parameters $\phi$.

In order to compute $\nabla_{\phi} u_j(\phi)$ we use the differentiable convex programming paradigm \cite{agrawal2019differentiable}. The DCP framework allows us to embed solutions to optimization problems as layers in neural networks, and also calculate gradients of output of the layers w.r.t. parameters in the optimization problem, so long as the problem formulation satisfies all the disciplined parameterized programming (DPP) properties.
The above convex optimization problem generally satisfies all the properties of disciplined parameterized programming (DPP) \cite{agrawal2019differentiable}, with parameter $\phi$, except for the objective function. Specifically, the constraints are all convex and are not functions of parameters. The objective function is also linear in variables $z^o_{ij}$, but should be an affine function of the parameters $\phi$ in order to satisfy the DPP constraints. Hence, we use the affine distance function $D(a,b) = |a-b|$. Hence, the DCP framework allows us to calculate $\nabla_{\phi} u_j(\phi) = \nabla_{z_{ij}^n} u_j \nabla_{\phi} z_{ij}^n(\phi)$. Algorithm \ref{algo:submodular} summarizes the proposed framework for jointly learning a deep learning model (parameterized by $\theta$) and a selection function for selecting a fraction $\gamma$ of the training examples, parameterized by $\phi$.

\textbf{Computational complexity:} Running time of the proposed algorithm scales linearly with the training dataset size. The algorithm also works at a time with a minibatch of training data, and hence is memory efficient. These two properties make the current algorithm work on very large training datasets, and complex deep learning models, provided there is enough computational power for inference of the model. A significant overhead of the current algorithm comes from solving an optimization problem for each parameter update. However, this cost can be controlled by keeping the minibatch sizes $|D_i|$ small, typically $|D_i|\leq 100$, which results in a linear optimization in 10000 variables. Another practical trick is to use a fixed number of datapoints for the old set $\cO(t)$, where the datapoints can be chosen to be most similar to points in the minibatch $D_i$ using a locality sensitive hashing based nearest neighbor search algorithm. Experimental results show that our algorithm takes a comparable amount of time as our closest competitor DVRL \cite{yoon2020data}, while being much more accurate.


\begin{algorithm}
 \caption{\textbf{\textit{HOST-CP}}: High-value Online Subset selection of Training samples through differentiable Convex Programming}
 \label{algo:submodular}
 \begin{algorithmic}[1]
 
 \State \textbf{Input:}
 \State \hspace{2mm} Training dataset $\cD$ with minibatches $D_1,\dots,D_k$, Test dataset $\cD^t$, fraction of selected instances $\gamma$
 \State \textbf{Output:}
 \State \hspace{2mm} Model Parameter $\theta$, Embedding model for selection parameter $\phi$
 \State{\textbf{Algorithm:}}
 \State Initialize old set $\cO(0)\rightarrow \Phi$ (Null set)
 \State Initialize parameters $\theta^0$ and $\phi^0$ randomly.
 \hspace{2mm}\For{each timestep $t=1,\dots,T$}
\State Let $D_{i(t)}$ be the current minibatch. 
\State for each datapoint $x_i\in D_{i(t)}$ and $x_j\in \cO(t)$, calculate embeddings $h(x_i,\phi)$ and $h(x_j,\phi)$
\State for each pair ($i,j$) , $x_i\in\cO$ and $x_j\in D_{i(t)}$ calculate $d^o(x_i,x_j)$
\State for each pair ($i,j$) , $x_i\in D_{i(t)}$ and $x_j\in D_{i(t)}$ calculate $d^n(x_i,x_j)$
\State Calculate $z_{ij}^n$ and $z_{ij}^o$ by solving optimization problem in Equation \ref{eq:opt2}
\State Calculate $u_j = \frac{1}{\xi} \max\{ \sum_{x_i\in D_{i(t)}} z^n_{ij}, \xi \}$ for all $x_j\in D_{i(t)}$
\State Include the selected points in the old set $\cO(t+1)$ for which $u_j=1$
\State Calculate $\cL(\theta^t;\phi^t,D_{i(t)})$ on the selected set by forward propagation.
\State Calculate $\theta^{t+1} = \theta^t - \alpha \frac{1}{k} \nabla_{\theta} \cL(\theta; \phi^t, D_{i(t)})$ by backpropagation.
\State Calculate $\hat{\theta} = \theta^t - \alpha \nabla_{\theta} \cL(\theta;\phi^t,D_{i(t)})$ followed by $\cV(\phi;\theta',\cD,\cD^t)$ using Equation \ref{eq:defCV}, where $\nabla_{\phi} u_j(\phi)$ is calculated using DCP as described in Section \ref{sec:trainlayer}
\State Update the embedding model parameter as: $\phi^{t+1} = \phi^t - \beta \frac{1}{k} \nabla_{\phi} \cV(\phi;\theta^t,D_{i(t)},\cD^t)$
 \EndFor
 \end{algorithmic}
\end{algorithm}
\section{Experiment}
In this section, we provide our experimental setup by describing the datasets and the different data valuation baselines. In Section \ref{sec:valtd}, we show the effectiveness of our proposed method - High-value Online Subset selection of Training samples through diferentiable Convex Programming(\textbf{\textit{HOST-CP}}) over the baselines in terms of performance metric with addition and removal of high value data points. Next in Section \ref{sec:mislabel}, we show the use case of diagnosing mislabelled examples which, when fixed can improve the performance. Following that in Section \ref{sec:qual}, we provide an analysis of the quality of subsets obtained by one of the baselines, DVRL \cite{yoon2020data} and the proposed method. We also show a ranking evaluation of selected points and their correspondence with ground truth relevance in Section \ref{sec:rank}. Lastly, Section \ref{sec:time} describes the performance of the proposed method in terms of its computational complexities.
\subsection{Experimental Setup}
\textbf{Dataset:} We have considered datasets of different modalities for our experiments. We use four types of datasets: CIFAR10 - a multi-class dataset for image classification, protein dataset \footnote{\href{protein}{\url{https://www.csie.ntu.edu.tw/~cjlin/libsvmtools/datasets/}}} for multi-class protein classification, 20newsgroups \footnote{\href{newsgroups20}{\url{https://scikit-learn.org/0.19/datasets/twenty_newsgroups.html}}} for text classification and a synthetic dataset for binary classification. Following \cite{ghorbani2019data}, we generate 10 synthetic datasets, using third-order polynomial function.

For CIFAR10, we obtain the image features from a pre-trained ResNet-18 model and train a 2 layer neural network on the learned features. In case of synthetic data and protein data, we pass the datapoints through a 3 layer neural network, whereas in case of 20newsgroups, we pass the TF-IDF features of the documents through a 3 layer network for classification.\\
\textbf{Baselines:} We consider four state-of-the-art techniques as our baselines. Two of the baselines - Influence Function - IF \cite{koh2017understanding} and TracIn-CP \cite{pruthi2020estimating} use the influence function method for data valuation, one uses data shapley (DS) value \cite{ghorbani2019data} for selecting high value data points while the fourth one, DVRL \cite{yoon2020data} uses reinforcement technique while sampling, to rank the data points.

\subsection{Valuable training data}
\label{sec:valtd}
\begin{figure*}[h!]
    \centering
    \begin{subfigure}{0.23\columnwidth}
    \includegraphics[width=\textwidth,height=3cm]{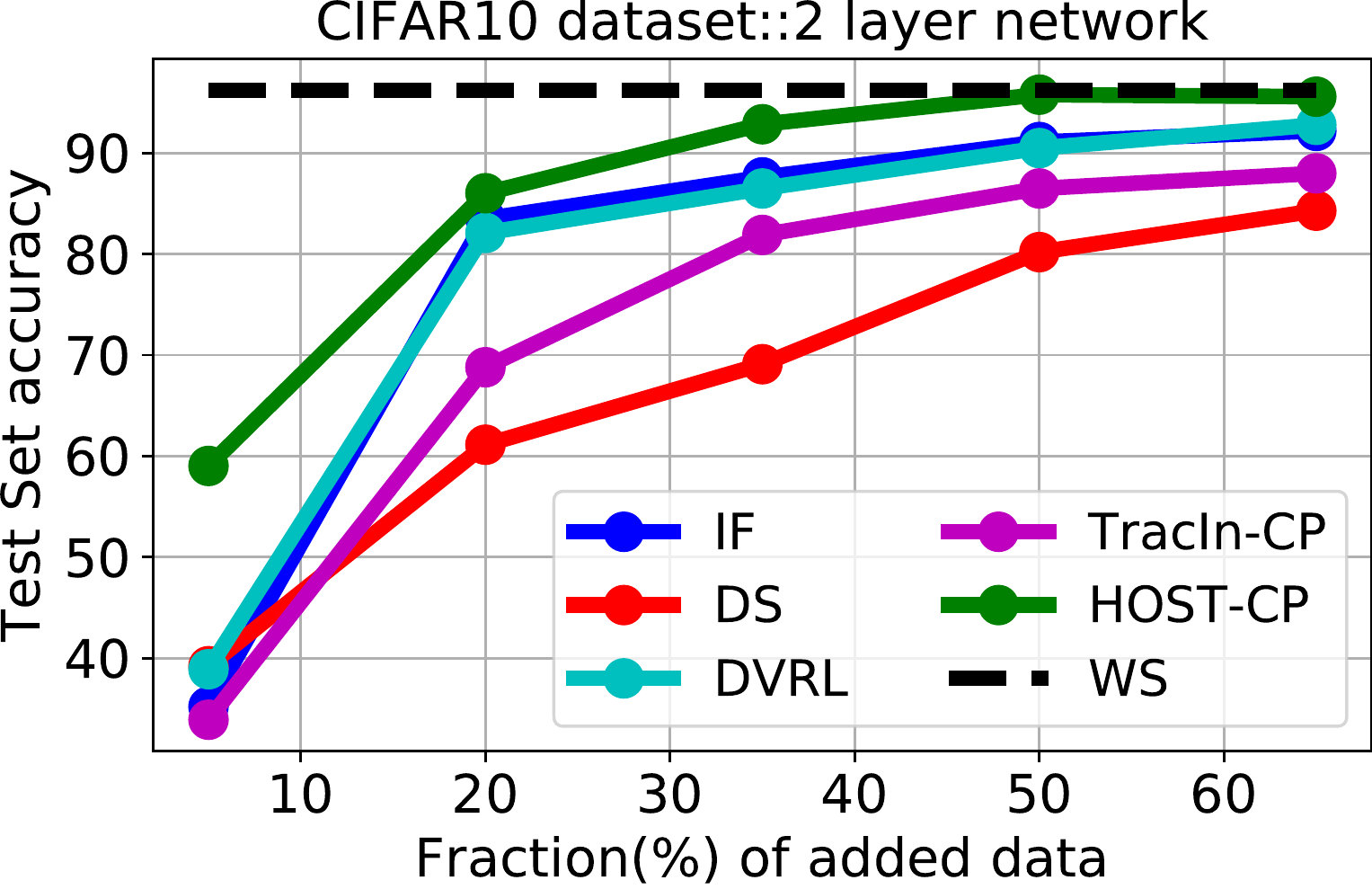}
    \end{subfigure}
    \begin{subfigure}{0.23\columnwidth}
    \includegraphics[width=\textwidth,height=3cm]{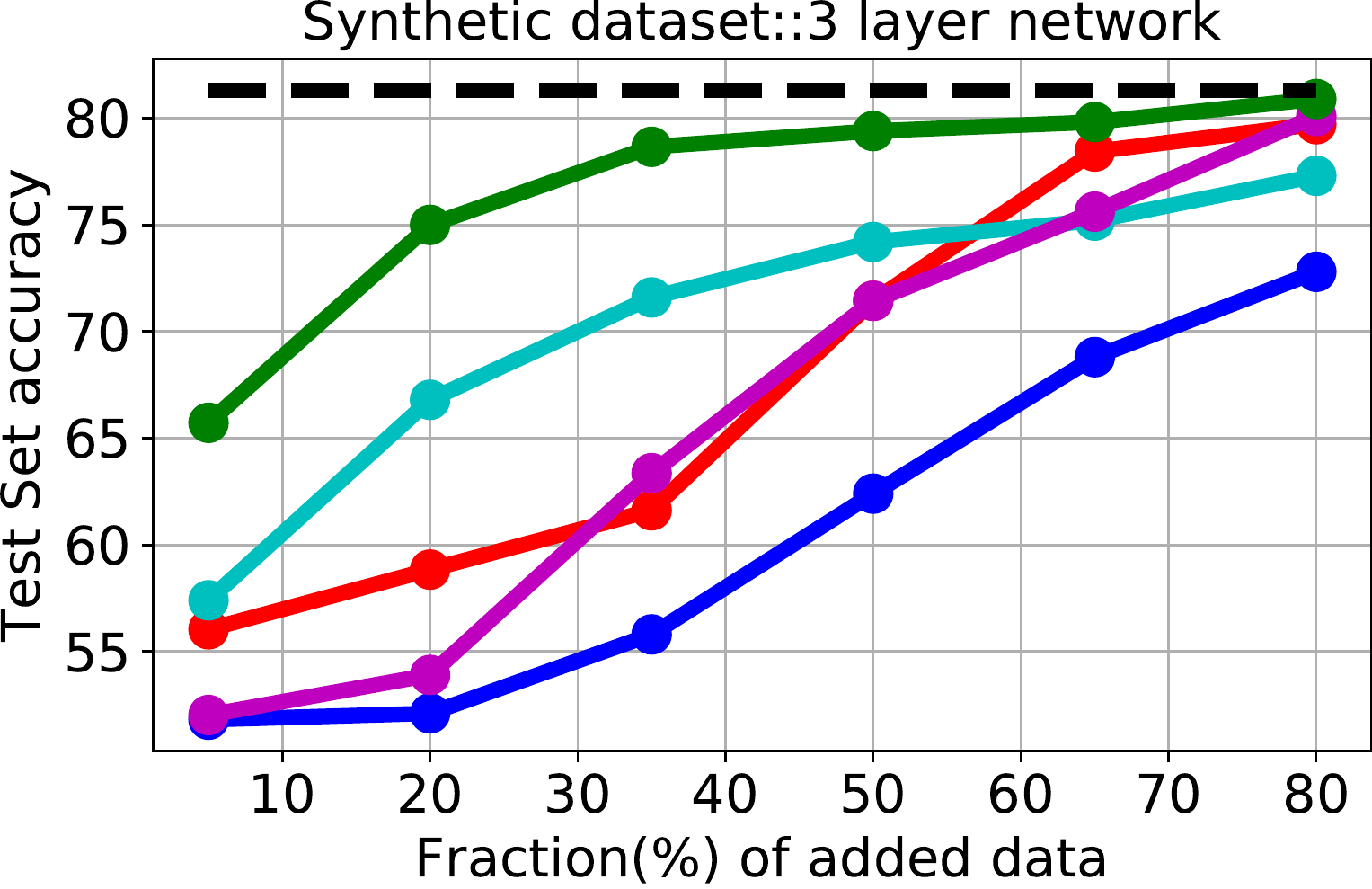}
    \end{subfigure}
    \begin{subfigure}{0.23\columnwidth}
    \includegraphics[width=\textwidth,height=3cm]{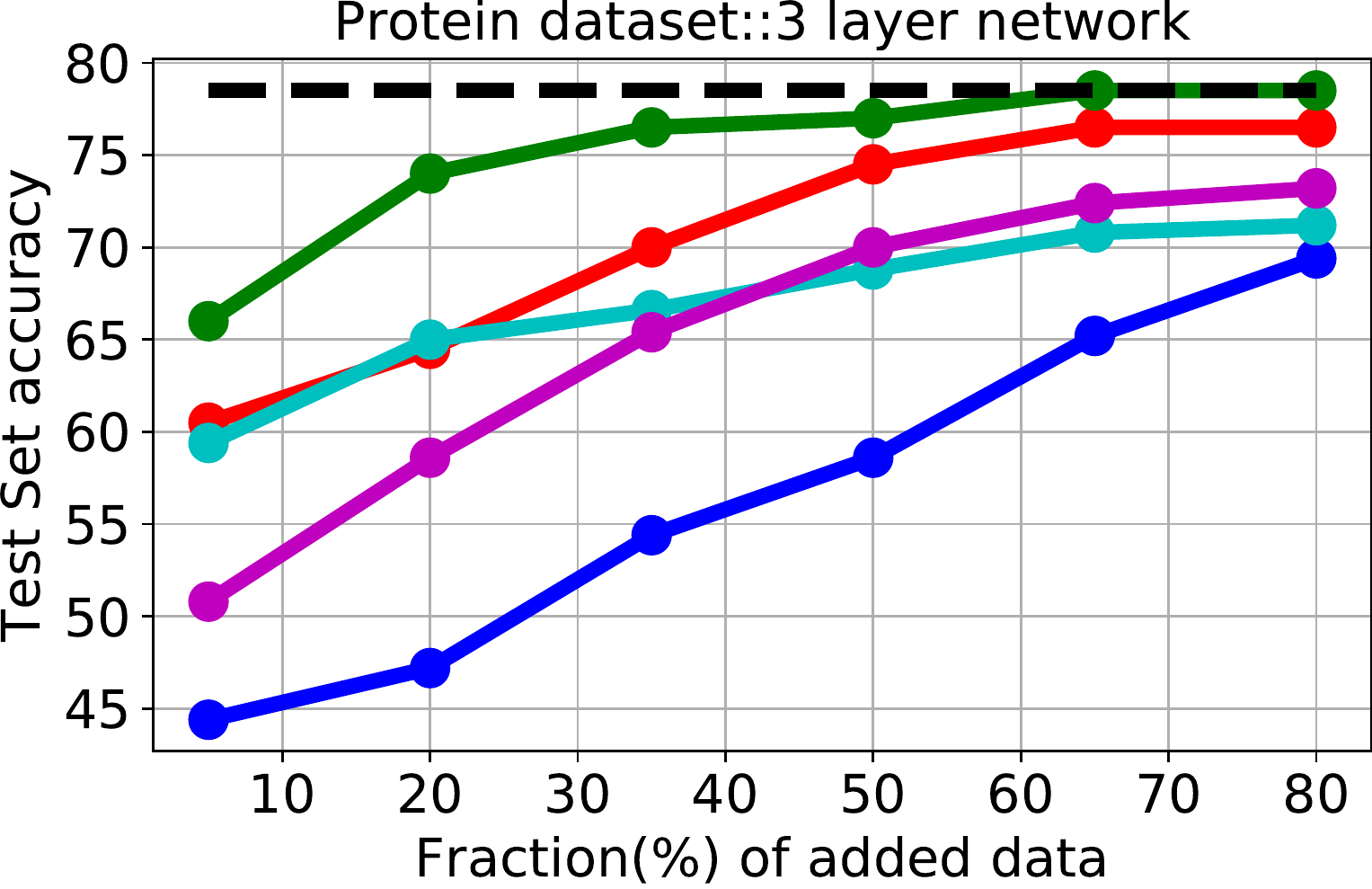}
    \end{subfigure}
    \begin{subfigure}{0.23\columnwidth}
    \includegraphics[width=\textwidth,height=3cm]{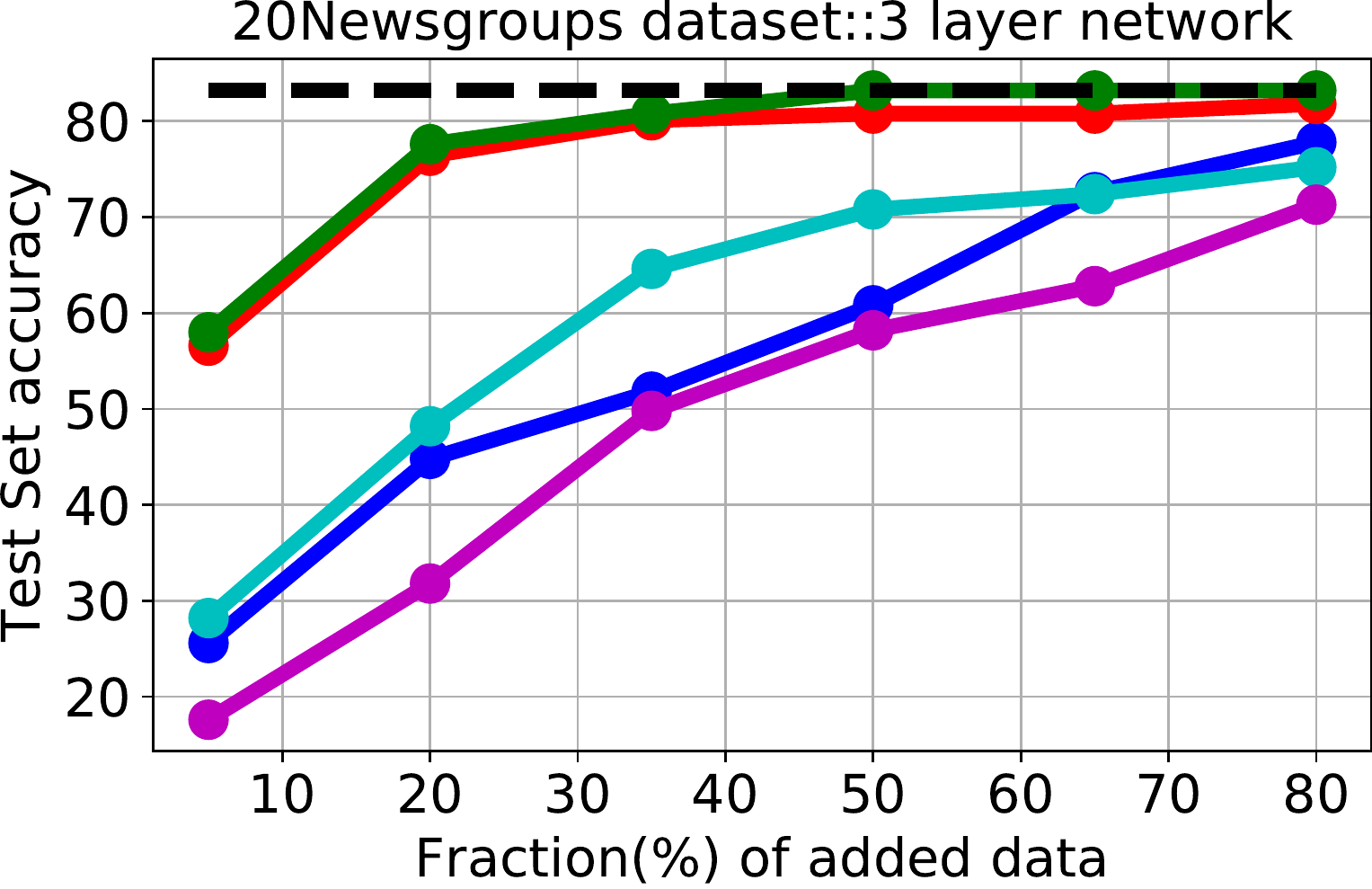}
    \end{subfigure}
    
    \begin{subfigure}{0.23\columnwidth}
    \includegraphics[width=\textwidth,height=3cm]{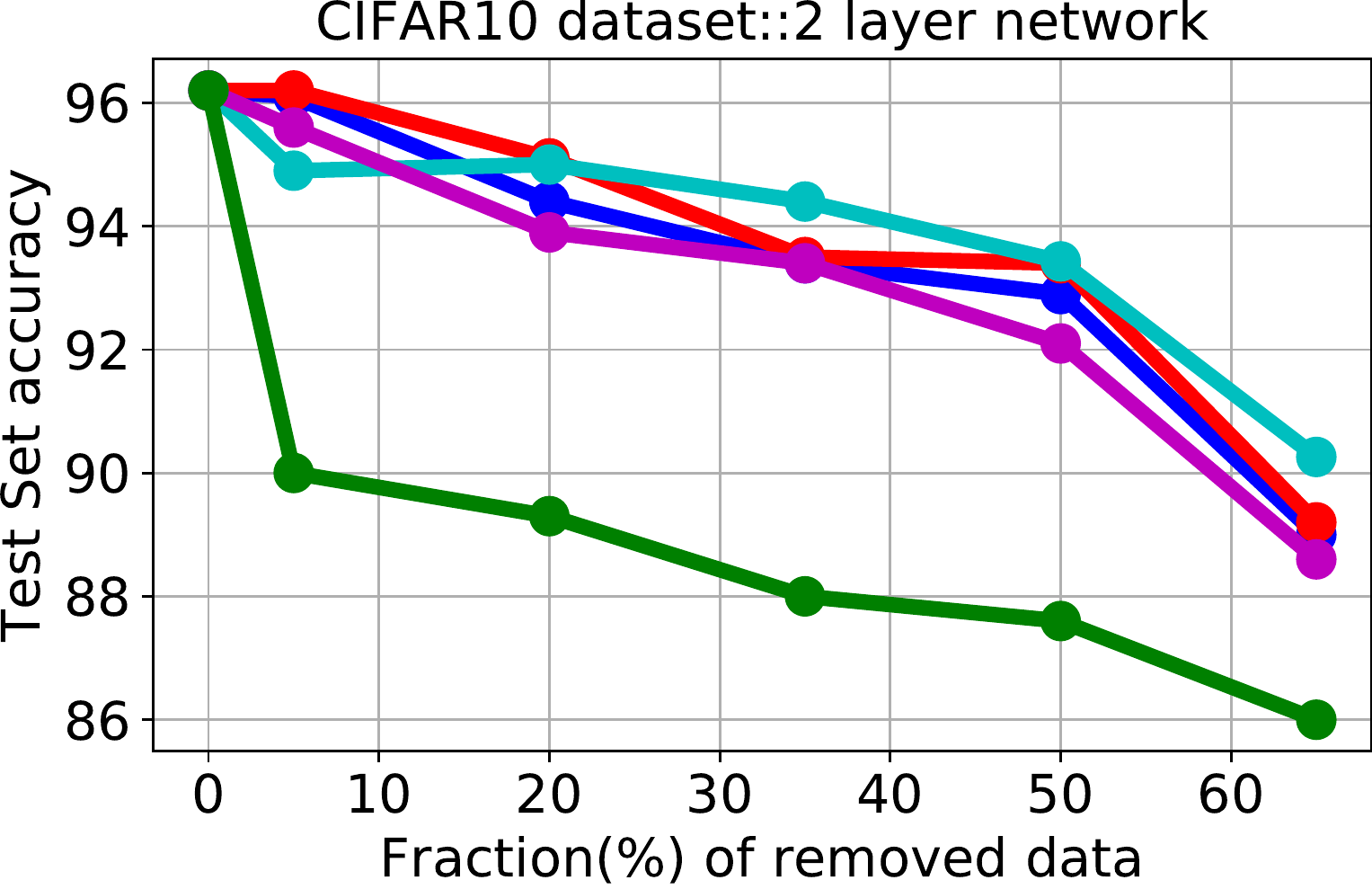}
    \end{subfigure}
    \begin{subfigure}{0.23\columnwidth}
    \includegraphics[width=\textwidth,height=3cm]{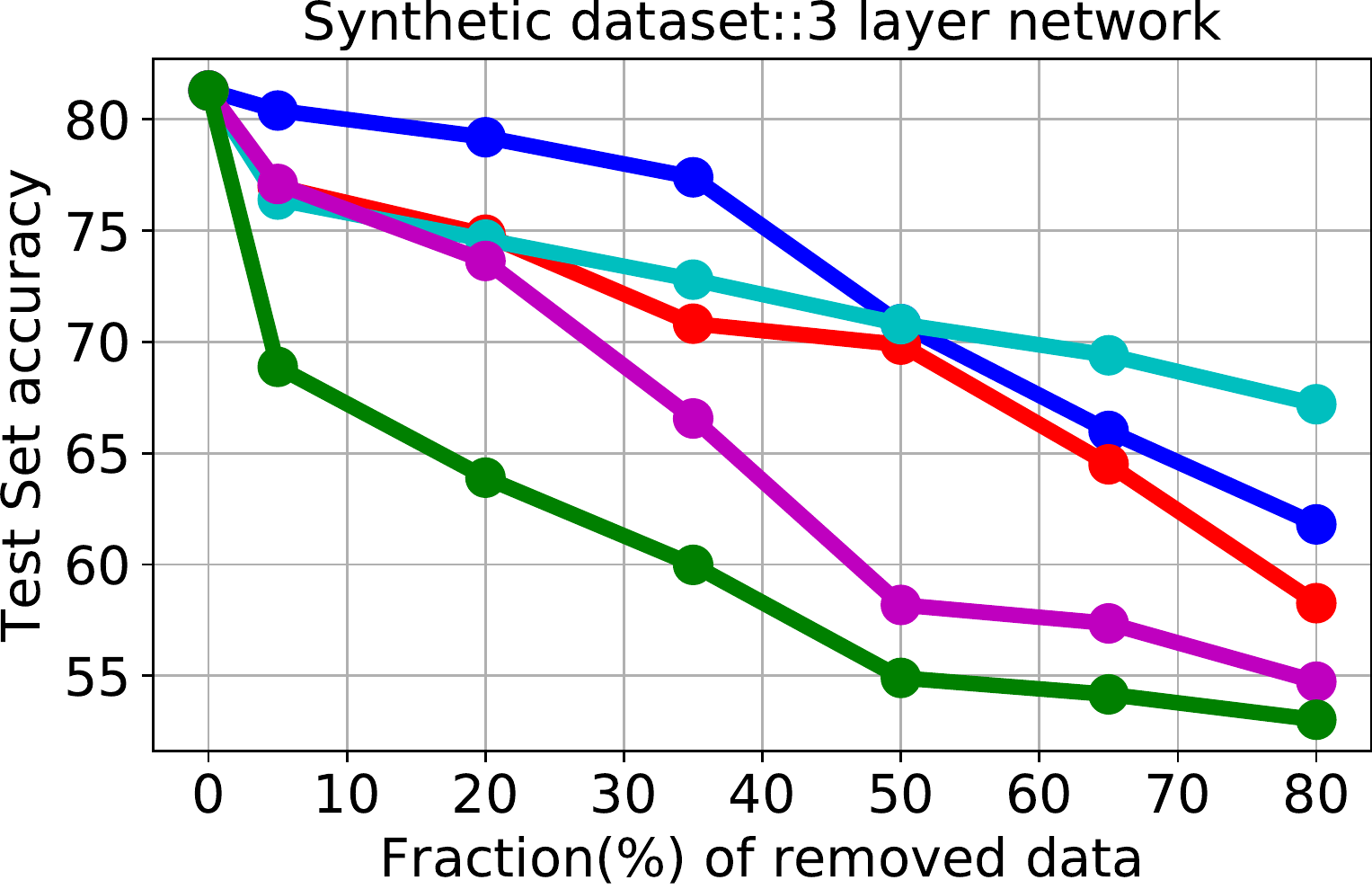}
    \end{subfigure}
    \begin{subfigure}{0.23\columnwidth}
    \includegraphics[width=\textwidth,height=3cm]{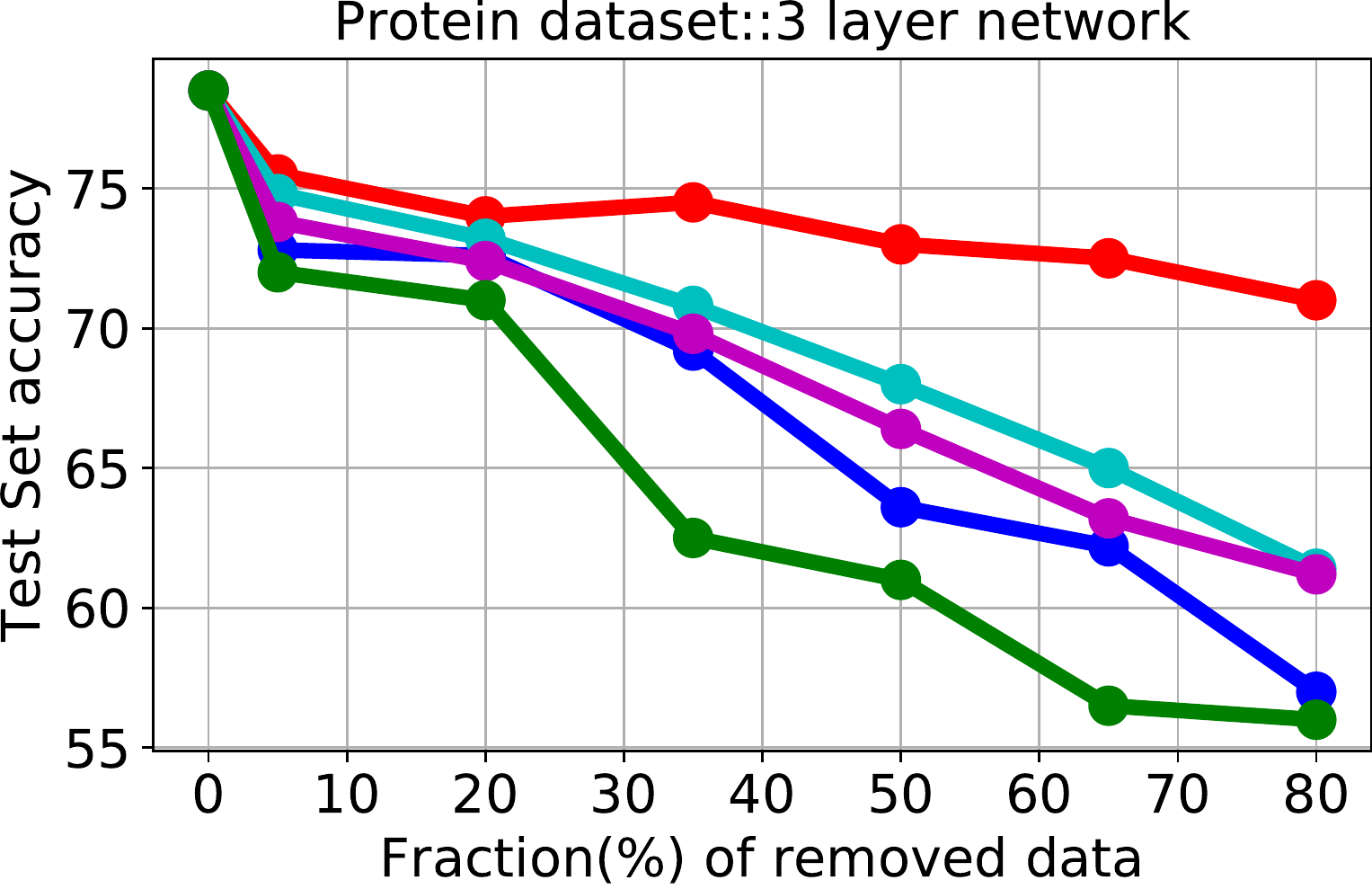}
    \end{subfigure}
    \begin{subfigure}{0.23\columnwidth}
    \includegraphics[width=\textwidth,height=3cm]{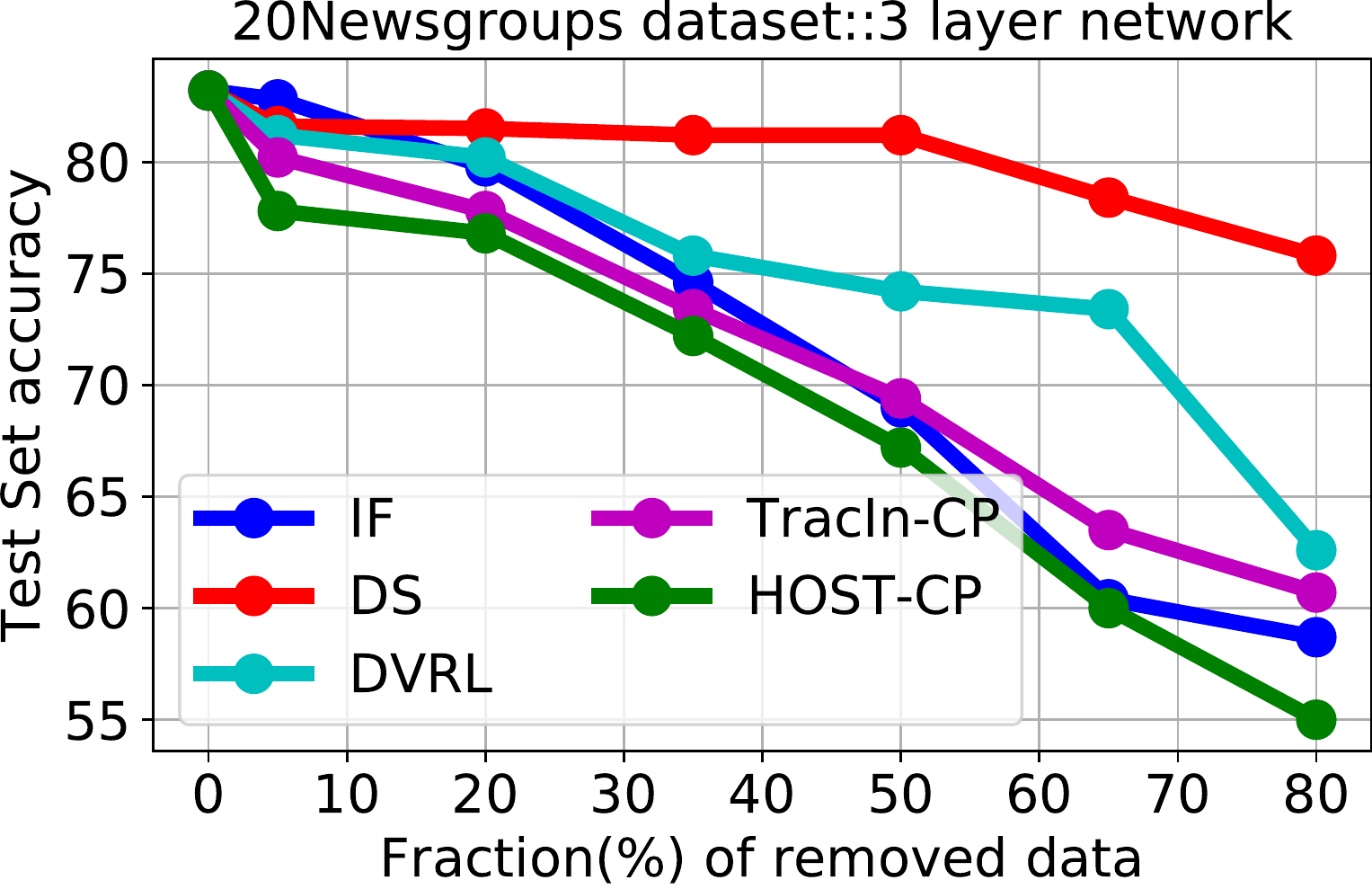}
    \end{subfigure}
    \caption{\textbf{Addition (top) and removal (bottom) of valuable training datapoints:} We compare HOST-CP with the aforementioned baselines (IF\cite{koh2017understanding}, DS\cite{ghorbani2019data}, TracIN-CP\cite{pruthi2020estimating}, DVRL\cite{yoon2020data}) in terms of performance metric (accuracy) on test set. 
    For each method, we select (for addition) or remove (for removal) top k\% important training points and assess their importance. The proposed method, HOST-CP performs consistently better(\textit{highest for addition and lowest for removal}) across all fractions and all datasets.}
    \label{fig:exp1}
\end{figure*}
We compare our method with state-of-the-art methods in terms of finding the most valuable training data points. We show the effectiveness of the proposed method by addition and removal of an incremental fraction of training instances to and from the dataset. We examine the test set performance for each fraction of selected or removed training points. The baseline methods are used to assign values to each data point, following which the top k\% fraction of data are selected/removed from the whole dataset to train the classification network models. Our proposed method has the flexibility to optimally select the significant k\% of data from the entire given dataset. We use the subsets of varying $k$ or $(100-k)$ fractions for training the networks and report the test set accuracies.

In Figure \ref{fig:exp1}, we report the test set accuracies after addition and removal of high-value training data points, that get selected using the respective methods. We can observe in Figure \ref{fig:exp1} \textit{(top)} that our method surpasses the baselines for all fractions and it approximately performs equivalent to the whole set (WS) within a fraction of 50 - 65\%. We also note a similar trend in removal of valuable data points. We can observe in Figure \ref{fig:exp1} \textit{(bottom)} that our method shows a significant drop in accuracy with removal of high-value data points. All the baselines show varying performances on different datasets. However, the proposed method consistently outperforms across all of them. This clearly suggests that HOST-CP is efficient enough to provide the optimal subset of training data points necessary for explaining the predictions on a given test set.
\subsection{Fixing mislabelled data}
\label{sec:mislabel}
Increase in data has led to rise in need of crowd sourcing for obtaining labels. Besides the chances of obtained labels being prone to errors \cite{frenay2013classification}, several data poisoning attacks are also being developed for mislabelling the data \cite{steinhardt2017certified}. In this experiment, we impute 10\% training data with incorrect labels, using which we compare our method with that of the baselines. The baseline methods consider that the mislabelled data shall lie towards the low ranked data points. We adopt the same idea and use the data points not selected by our method (reverse-selection) for analysis.

\begin{figure}[h!]
    \centering
    \begin{subfigure}{0.3\columnwidth}
    \includegraphics[width=\textwidth,height=4cm]{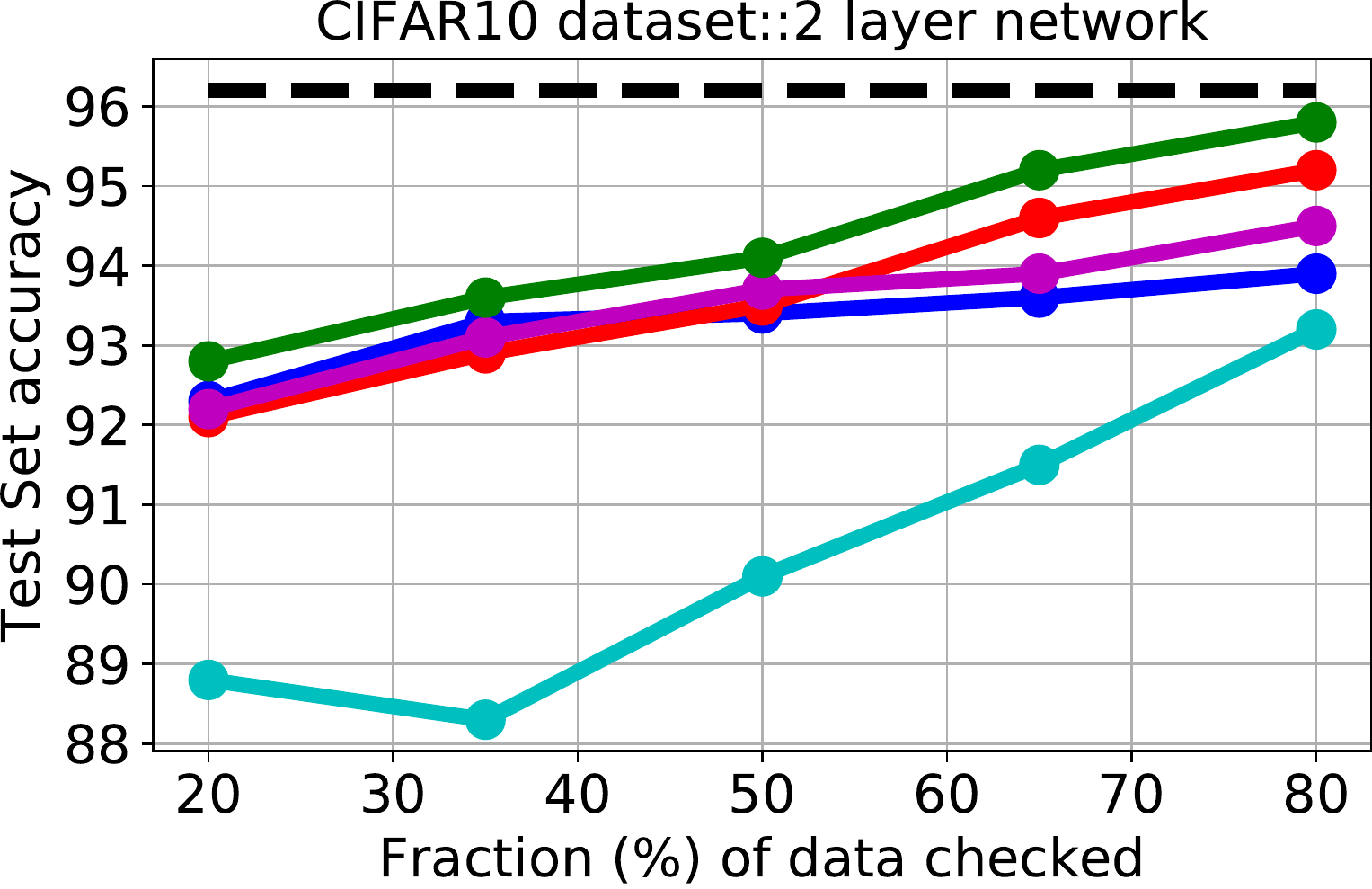}
    \end{subfigure}
    \begin{subfigure}{0.3\columnwidth}
    \includegraphics[width=\textwidth,height=4cm]{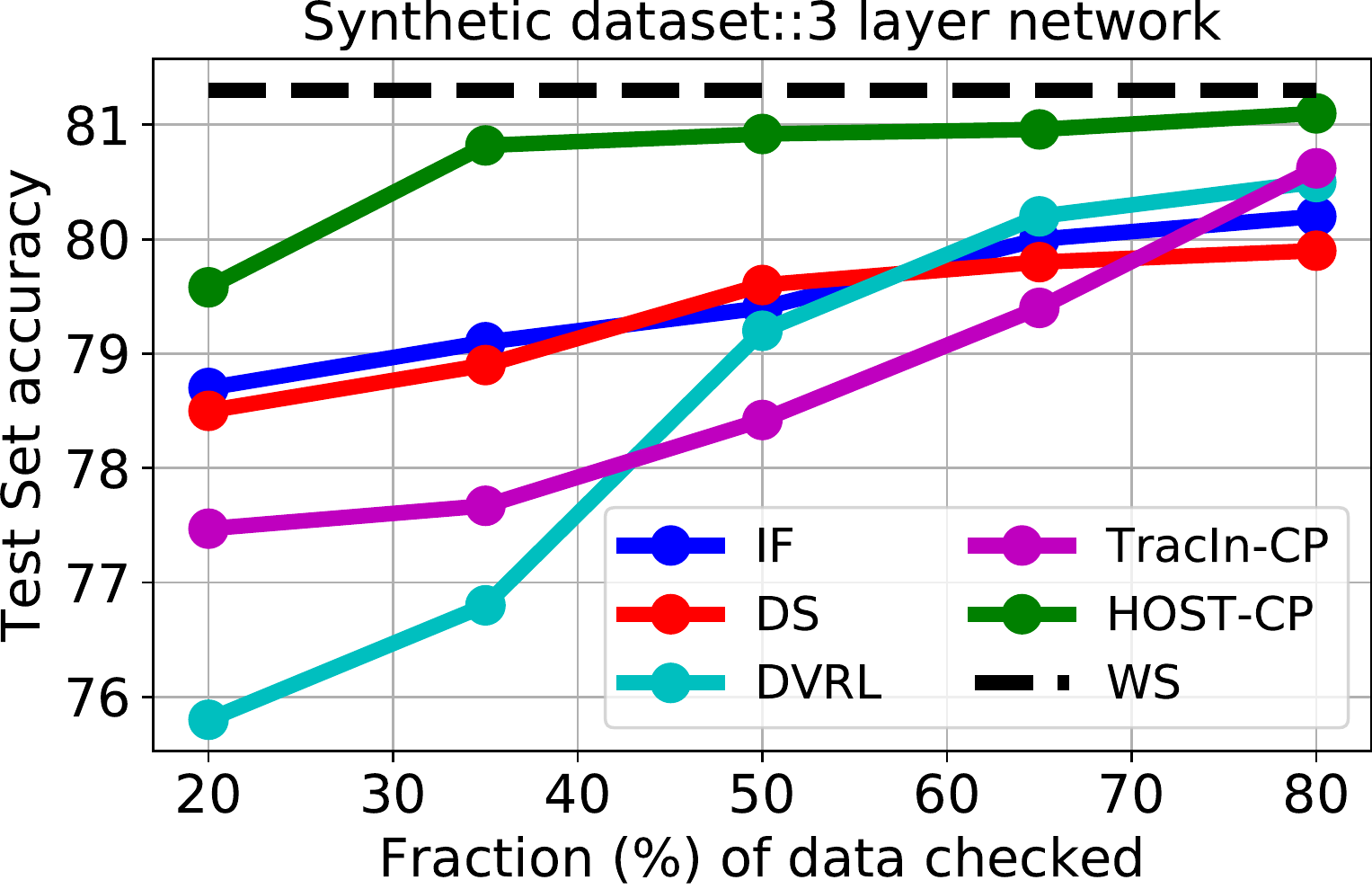}
    \end{subfigure}
    \caption{\textbf{Accuracy after fixing mislabelled data:} For every k\% fraction of inspected data, we fix the mislabelled instances and re-train. HOST-CP shows improved performance with each fixing of labels.}
    \label{fig:exp2}
\end{figure}

In this experiment, we inspect the bottom k\% fraction (for baselines) and unselected k\% fraction of the data (for the proposed method) for presence of wrong labels. 
We run this experiment on CIFAR10 and synthetic dataset and report the fraction of incorrect labels fixed with varying k. While imputing 10\% of training data, we flip the labels of synthetic dataset since it consists of binary labels, while in case of CIFAR10, we change the original label to one out of 9 classes. We can observe in Table \ref{tab:flipfix} that the proposed method is able to find comparable mislabelled fraction of incorrect labels in comparison to the baselines. We also run another experiment where, we fix the incorrect labels in the considered k\% fraction of data and train the entire data. Figure \ref{fig:exp2} shows a rise in test set accuracy with a significant margin using the proposed method. This denotes the effectiveness of HOST-CP, in diagnosing mislabelled examples which turns out to be an important use-case in data valuation techniques.

\begin{table}[]
\centering
\caption{\textbf{Diagnosing mislabelled data:} We inspect the training points starting from the least significant data and report the fraction of labels fixed from the inspected data. Overall, the proposed method, HOST-CP detects a higher rate of mislabelled data.}
\label{tab:flipfix}
\begin{tabular}{|l|l|l|l|l|l|l|l|l|l|l|}
\hline
\multicolumn{1}{|c|}{\multirow{3}{*}{\textbf{\begin{tabular}[c]{@{}c@{}}Fraction of\\ data checked\\ (\%)\end{tabular}}}} & \multicolumn{10}{c|}{\textbf{Fraction of incorrect labels fixed}}                                                                                                                                                                                                                                                                                                                                                                                                                                                                                               \\ \cline{2-11} 
\multicolumn{1}{|c|}{}                                                                                                    & \multicolumn{5}{c|}{\textbf{CIFAR10}}                                                                                                                                                                                                                                          & \multicolumn{5}{c|}{\textbf{Synthetic Data}}                                                                                                                                                                                                                                   \\ \cline{2-11} 
\multicolumn{1}{|c|}{}                                                                                                    & \multicolumn{1}{c|}{\textbf{IF}} & \multicolumn{1}{c|}{\textbf{DS}} & \multicolumn{1}{c|}{\textbf{DVRL}} & \multicolumn{1}{c|}{\textbf{\begin{tabular}[c]{@{}c@{}}TracIn\\ -CP\end{tabular}}} & \multicolumn{1}{c|}{\textbf{\begin{tabular}[c]{@{}c@{}}HOST\\ -CP\end{tabular}}} & \multicolumn{1}{c|}{\textbf{IF}} & \multicolumn{1}{c|}{\textbf{DS}} & \multicolumn{1}{c|}{\textbf{DVRL}} & \multicolumn{1}{c|}{\textbf{\begin{tabular}[c]{@{}c@{}}TracIn\\ -CP\end{tabular}}} & \multicolumn{1}{c|}{\textbf{\begin{tabular}[c]{@{}c@{}}HOST\\ -CP\end{tabular}}} \\ \hline
20                                                                                                                        & 0.21                             & 0.19                             & 0.195                                   & 0.20                                                                                   & \textbf{0.23}                                                                  & 0.22                             & 0.22                             & 0.19                                   & 0.23                                                                                   & 0.23                                                                           \\ \hline
35                                                                                                                        & 0.35                             & 0.34                             &  0.336                                  & 0.35                                                                                   & \textbf{0.36}                                                                  & 0.34                             & 0.37                             & 0.33                                   & 0.35                                                                                  & \textbf{0.39}                                                                  \\ \hline
50                                                                                                                        & 0.50                             & 0.50                             & 0.49                                   & 0.50                                                                                   & \textbf{0.51}                                                                  & 0.43                             & 0.54                             & 0.46                                   & 0.51                                                                                   & \textbf{0.55}                                                                  \\ \hline
65                                                                                                                        & 0.61                             & 0.61                             & 0.65                                   & 0.63                                                                                   & \textbf{0.67}                                                                  & 0.56                             & 0.67                             & 0.58                                   & 0.68                                                                                   & \textbf{0.68}                                                                  \\ \hline
80                                                                                                                        & 0.74                             & 0.78                             & 0.80                                   & 0.79                                                                                   & \textbf{0.81}                                                                  & 0.66                             & 0.81                             & 0.79                                   & 0.81                                                                                   & \textbf{0.83}                                                                  \\ \hline
\end{tabular}
\end{table}

\subsection{Qualitative analogy of data valuation}
\label{sec:qual}
\begin{figure}[h!]
    \centering
    
    \includegraphics[width=0.5\textwidth,height=4.2cm]{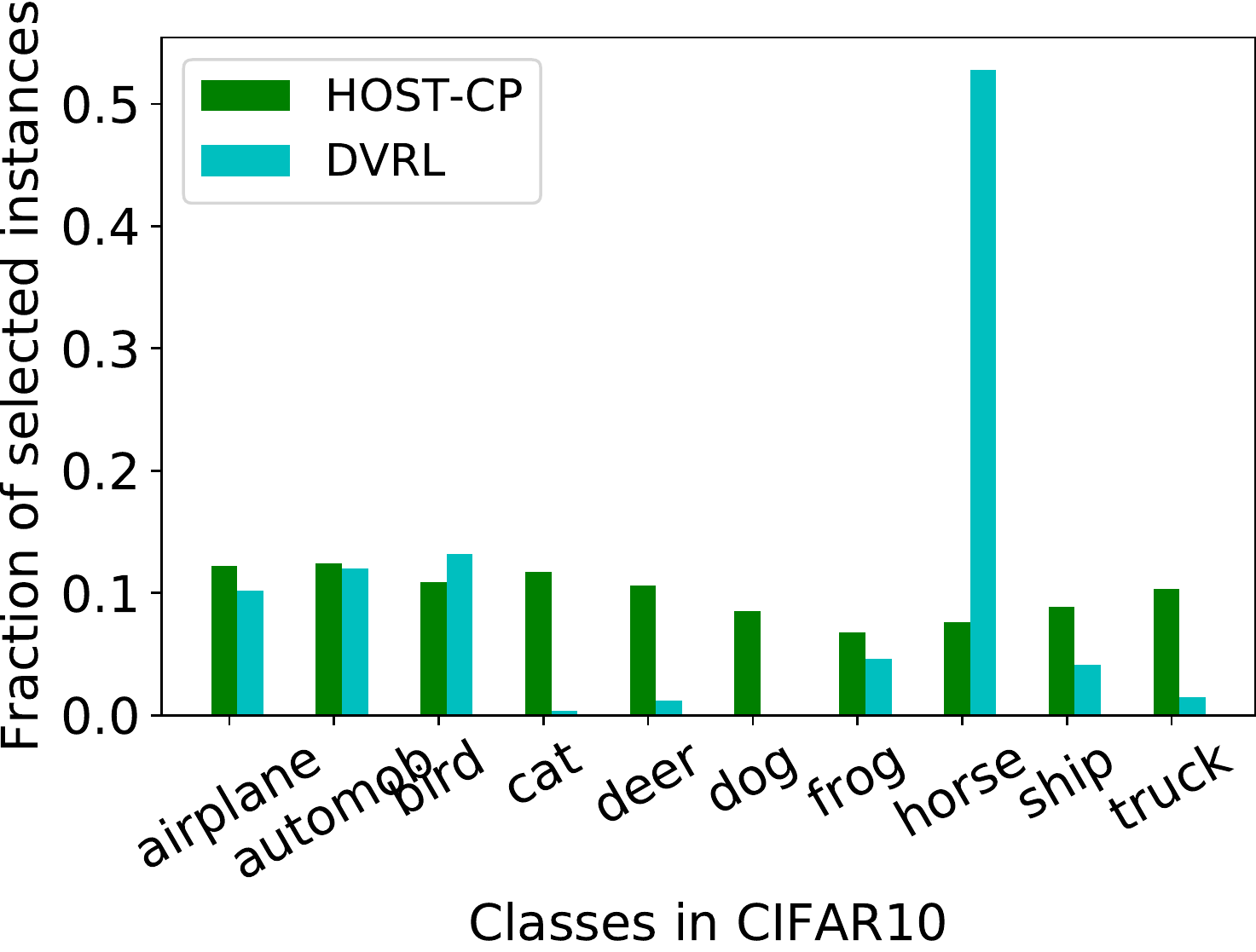}
    \caption{\textbf{Quality of subset:} We observe that unlike DVRL, HOST-CP selects diverse images across the classes leading to better performance.}
    \label{fig:analysis}
\end{figure}

As we had mentioned earlier, methodologically, DVRL \cite{yoon2020data} is the closest method to our approach. Hence, in order to qualitatively explain the difference in performance between that of DVRL and the proposed method, we try to analyse the quality of subsets obtained by both of them on CIFAR10 dataset. Figure \ref{fig:analysis} shows the fraction of selected images across classes by both the methods. We compare the top 5\% selected subset by DVRL and the proposed method in terms of class distribution. Unlike the proposed method which compares the past selected data points with the incoming points to compute the new subset, DVRL keeps track of past average test losses and values the instances based on the marginal difference with the current loss. This leads to a certain bias in case of DVRL, towards some particular class of images or similar high-performing datapoints which are more contributory towards reduction in test set losses. Following this, we observe that HOST-CP selects a far more diverse set of samples, thus justifying its better performance.

\subsection{Ranking function for value points}
\label{sec:rank}
\begin{figure*}[h!]
    \centering
    \begin{subfigure}{0.25\columnwidth}
    \includegraphics[width=\textwidth,height=3cm]{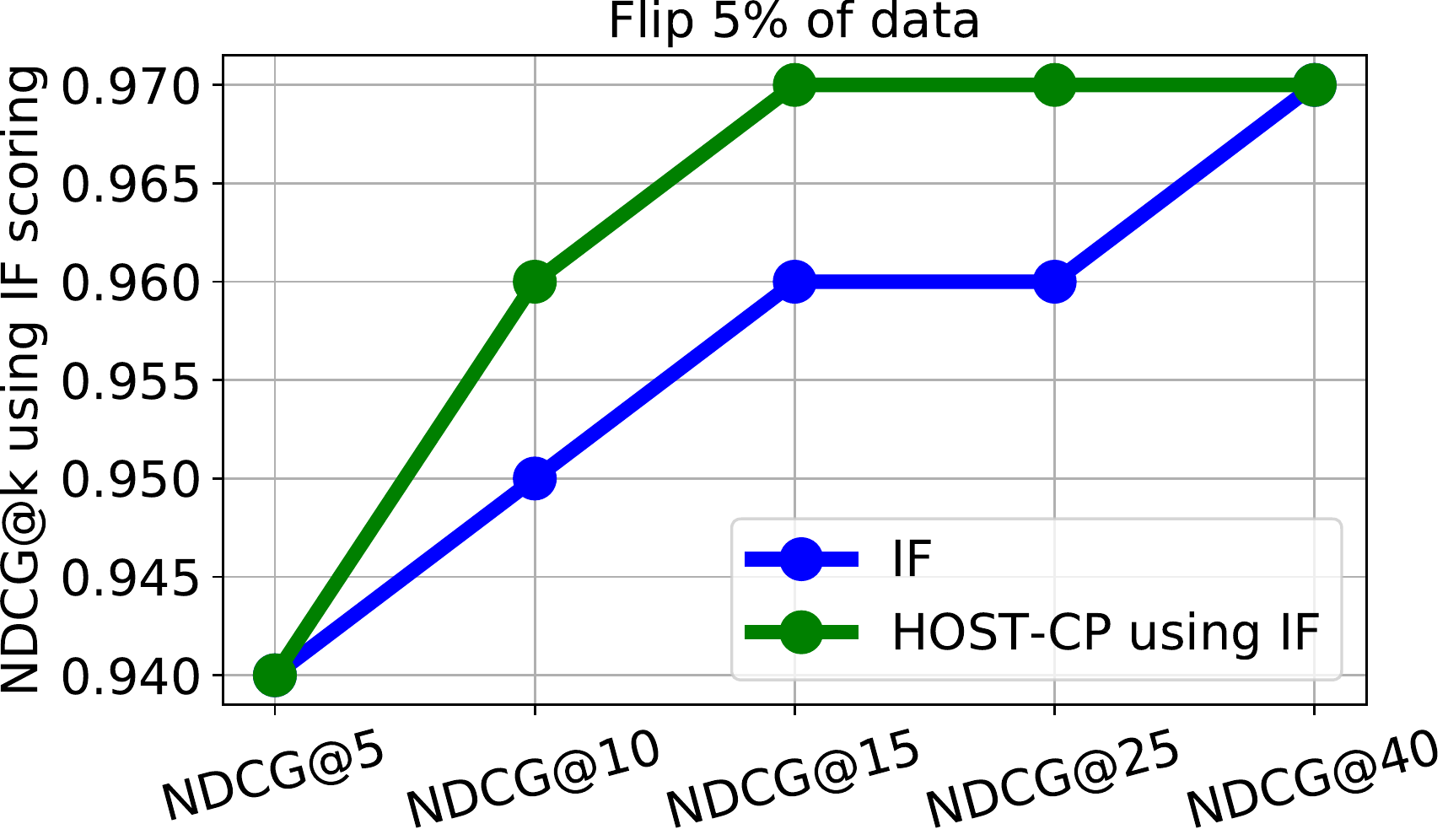}
    \end{subfigure}
    \begin{subfigure}{0.25\columnwidth}
    \includegraphics[width=\textwidth,height=3cm]{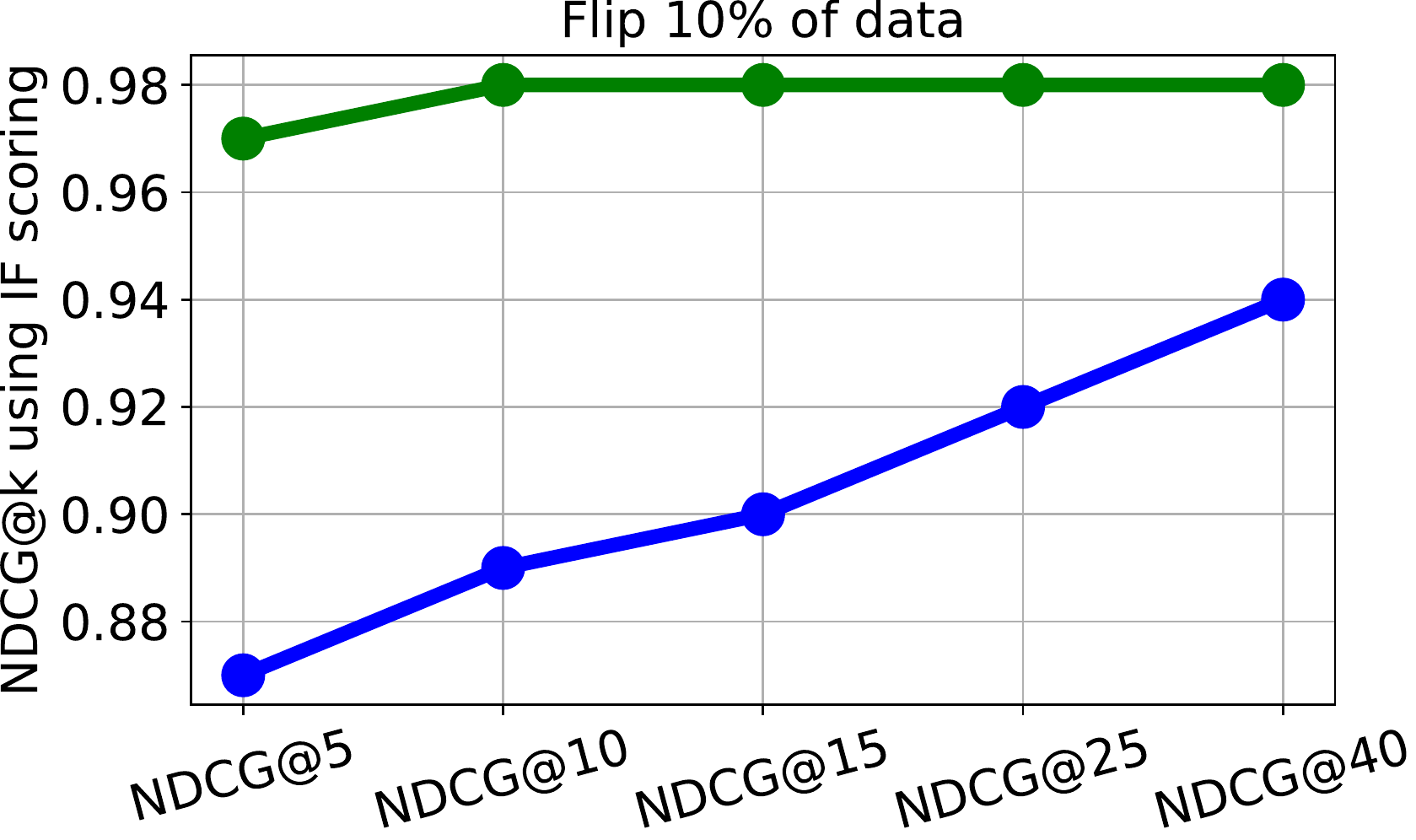}
    \end{subfigure}
    \begin{subfigure}{0.25\columnwidth}
    \includegraphics[width=\textwidth,height=3cm]{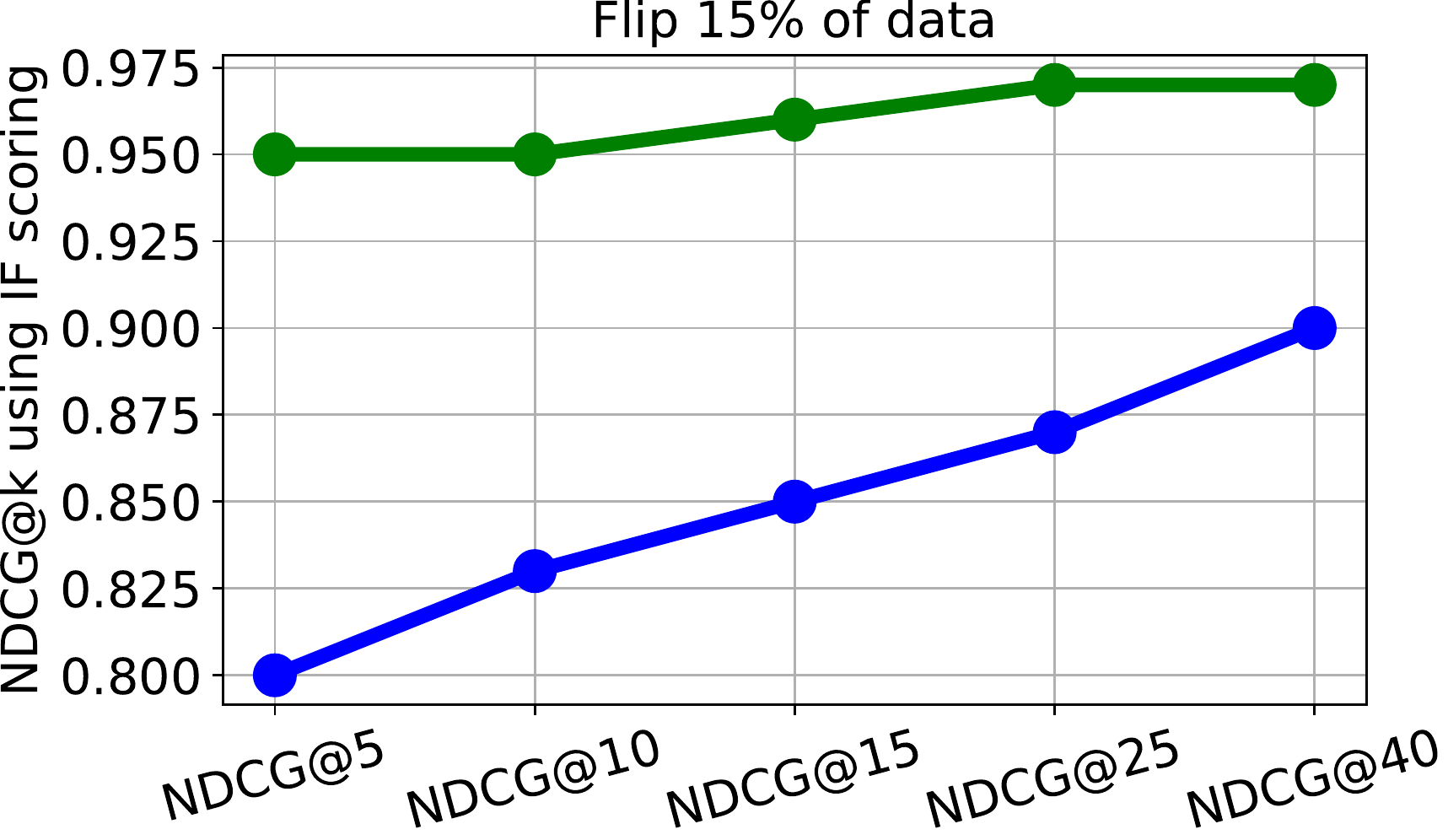}
    \end{subfigure}
    
    \begin{subfigure}{0.25\columnwidth}
    \includegraphics[width=\textwidth,height=3cm]{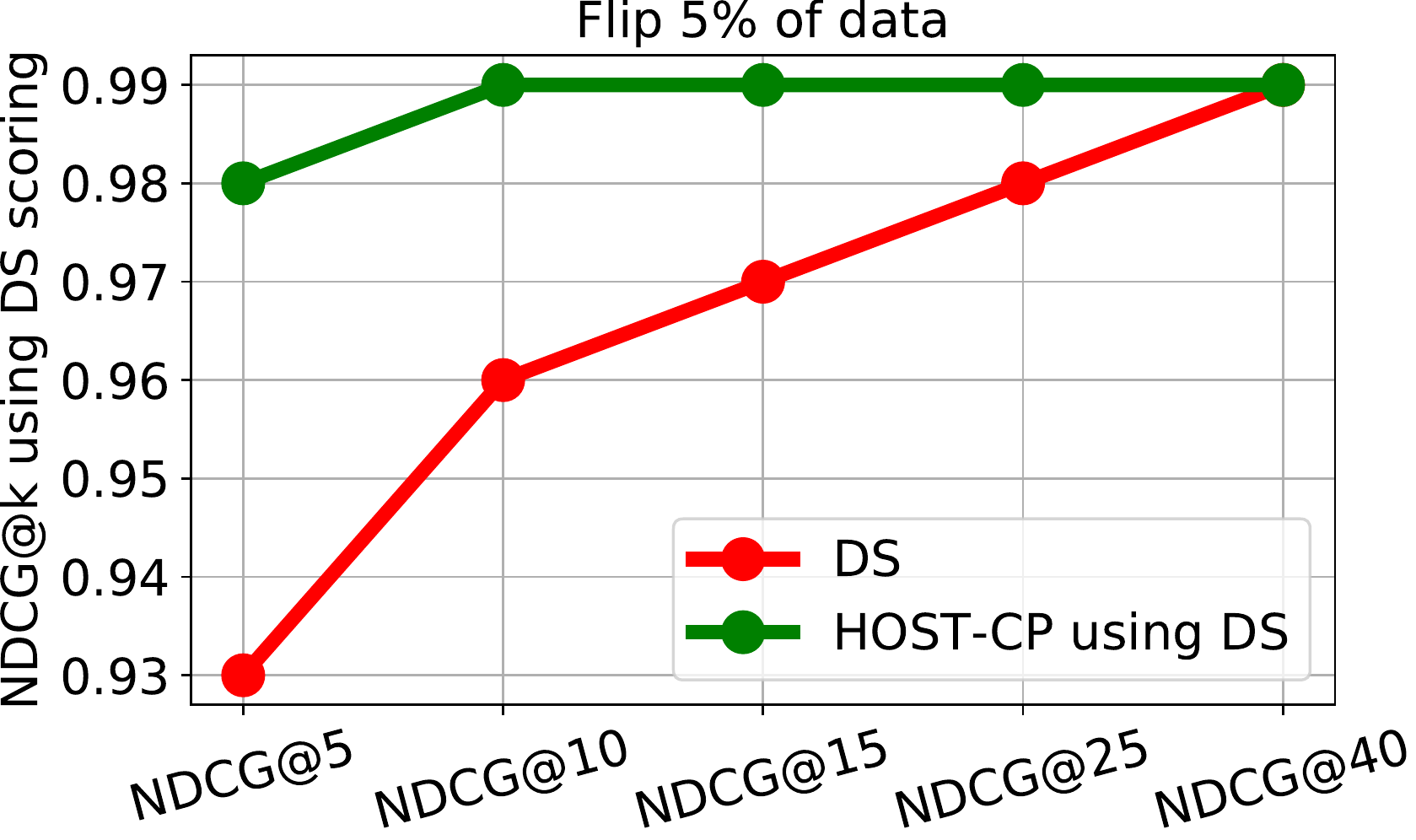}
    \end{subfigure}
     \begin{subfigure}{0.25\columnwidth}
    \includegraphics[width=\textwidth,height=3cm]{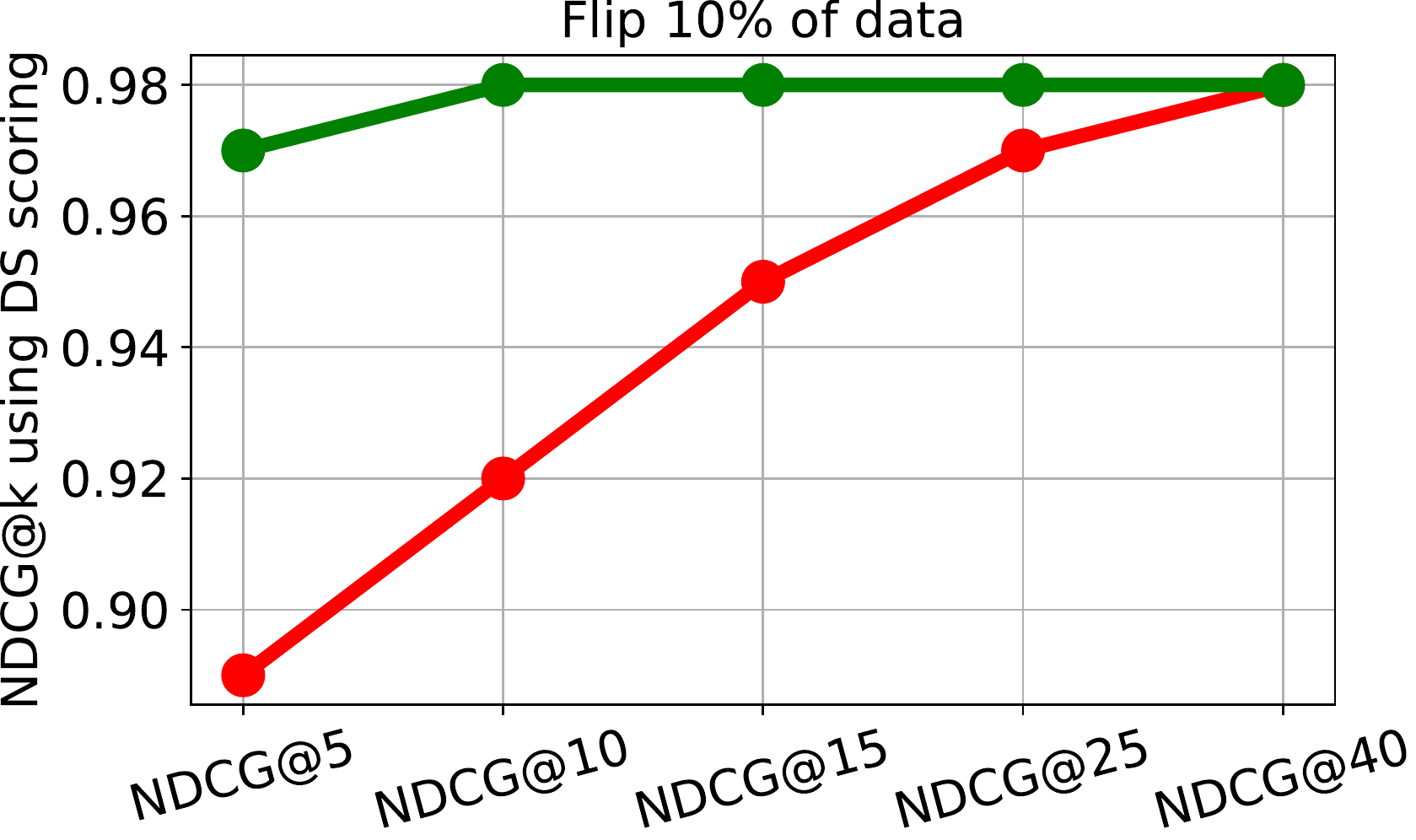}
    \end{subfigure}
    \begin{subfigure}{0.25\columnwidth}
    \includegraphics[width=\textwidth,height=3cm]{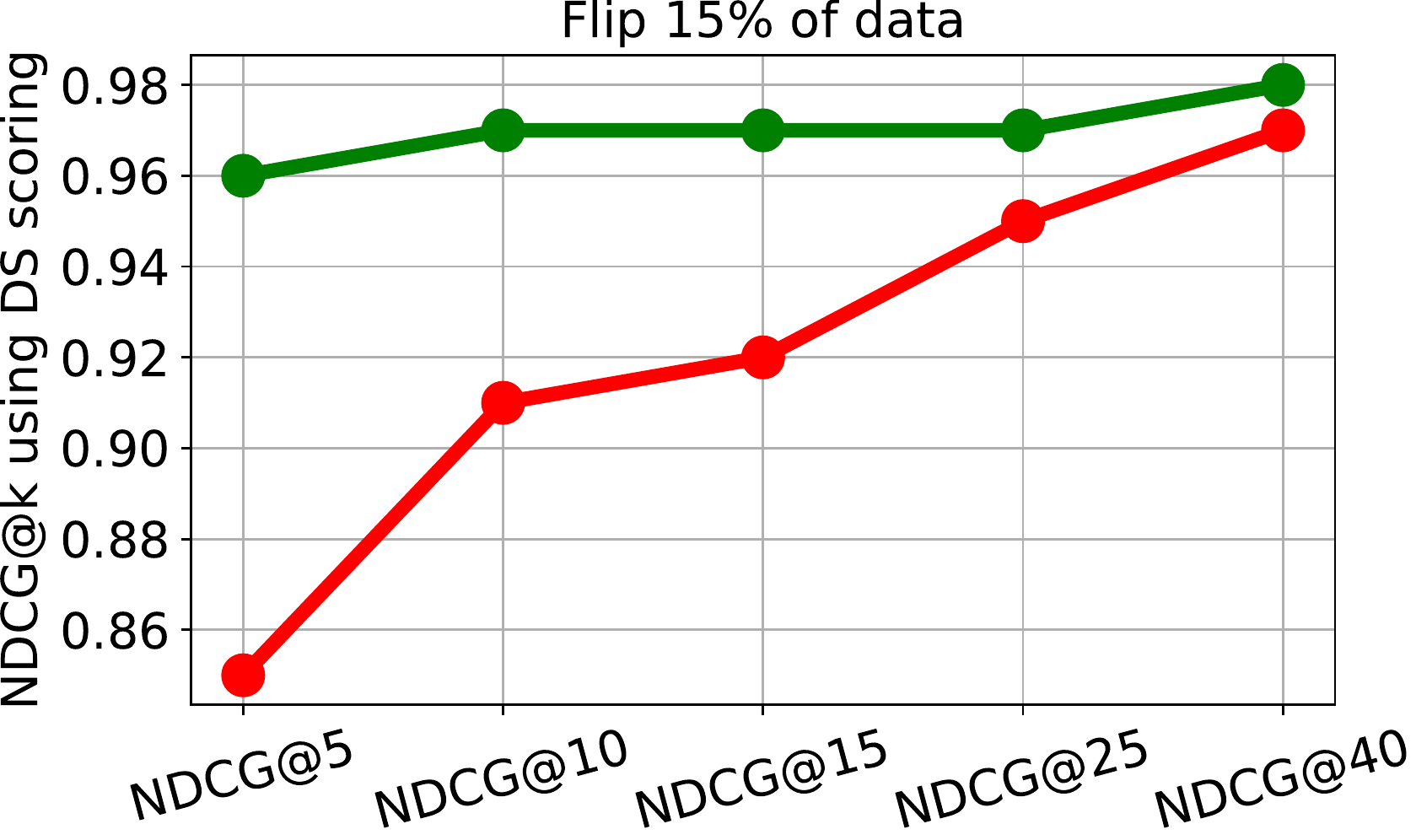}
    \end{subfigure}
    \caption{\textbf{Ranking of data points:} We select k\% fraction of data for each method and provide scores for each datum. While IF and DS subsets inherently come along with scores, the proposed method, HOST-CP returns a subset which is scored using IF and DS. Following the acquisition of scores for each data point, we use NDCG@k to calculate the ranking values. The proposed method is found to have a higher NDCG score.}
    \label{fig:rank}
\end{figure*}
Unlike the baseline methods which provide a score to each data point, the proposed method has no such analogous scoring function. It is designed to return the appropriate explainable subset necessary for optimum performance on a given test set. We design this experiment to assess the significance of the selected data points in terms of ranking function values. We use NDCG score for reporting the ranking value.

We use the synthetic dataset and create a series of ground-truth values by flipping n\% \{=5,10,15\} of the data. We assign ground-truth to the flipped and unflipped points with 0 and 1 respectively. We examine the bottom(or low-value) k\% of the data for computing the ranking values since the flipped points are more inclined to lie towards the bottom. As we have observed in Section \ref{sec:mislabel}, we use the reverse-selection procedure to obtain the bottom k\% of data for the proposed method, while for the baselines, the low-valued k\% points occupy the bottom rung.

In order to compute NDCG scores, we adopt the two baselines \cite{koh2017understanding},\cite{ghorbani2019data} to provide scores to the subset of points obtained using the proposed method. In case of the baselines, we use the k-lowest scoring points for computing the rank values.

We flip 5\%, 10\% and 15\% of the data and compute NDCG@k for k $\in$ \{5,10,15,25,40\}. Here $k$ refers to the bottom $k\%$ subset from the methods. In Figure \ref{fig:rank}, we compare the respective ranking values(NDCG) of subsets obtained using influence function (IF) or data shapley (DS) with the subset obtained using the proposed method followed by usage of the corresponding scoring method (IF or DS) on this acquired subset. We can observe that NDCG values obtained by the proposed method's subset followed by scoring using IF or DS always stays ahead than using the subsets obtained by IF or DS, starting from an early fraction. This shows that the subset obtained using the proposed method, HOST-CP is efficient enough in keeping the proportions of relevant(unflipped) and non-relevant(flipped) points in their respective positions leading to a higher-NDCG score.


\subsection{Computational complexity}
\label{sec:time}
\begin{figure}[h!]
    \centering
    \begin{subfigure}{0.5\columnwidth}
    \includegraphics[width=\textwidth,height=4cm]{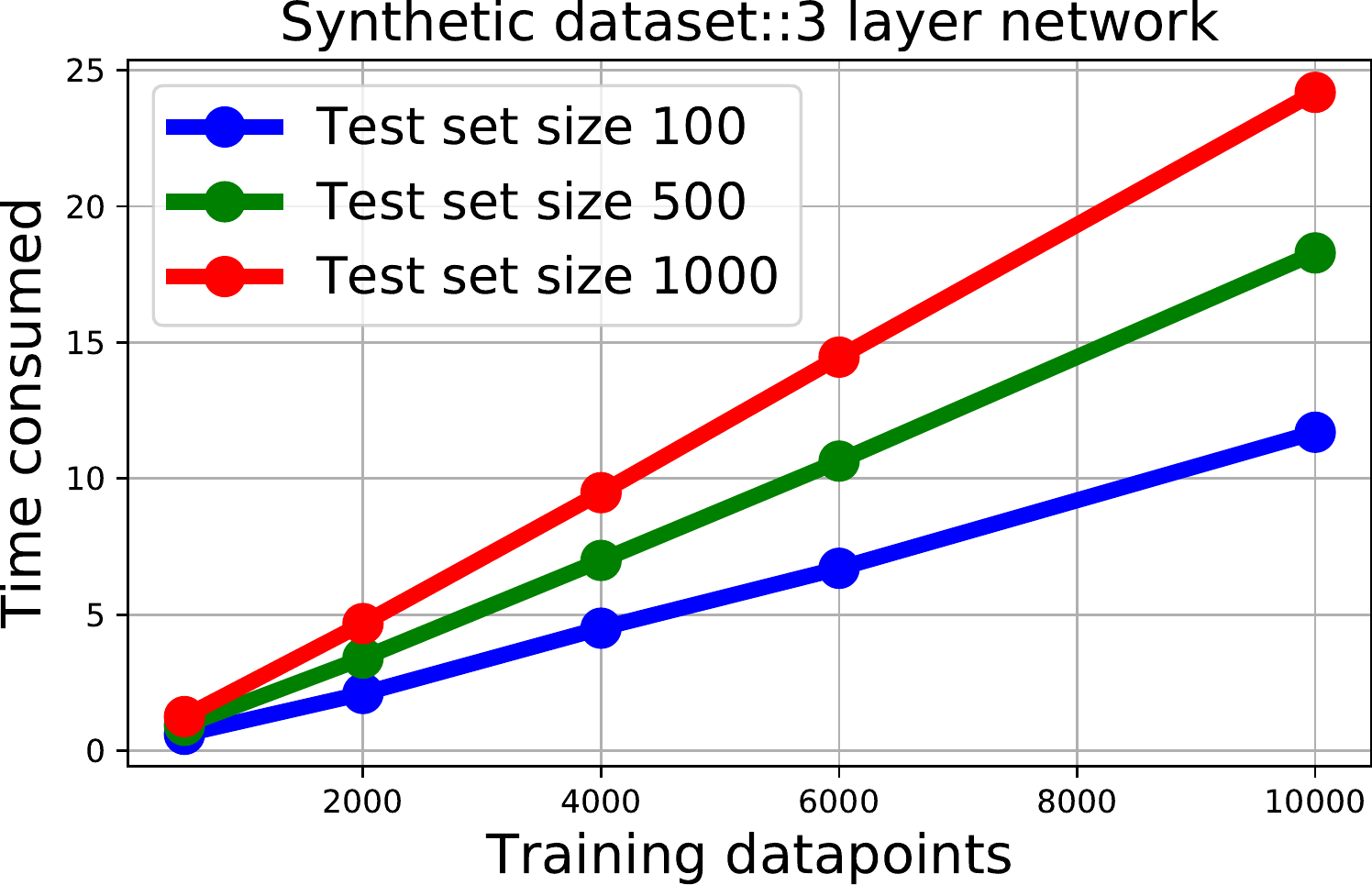}
    \end{subfigure}
    \caption{\textbf{Scaling with data:} We show that the change in time consumed by HOST-CP with variation in training data points and test data points using a fixed network is linear with increasing data.}
    \label{fig:timec}
\end{figure}

\begin{table}[]
\centering
\caption{\textbf{Time comparison}: We compare the times consumed by DVRL and HOST-CP. We observe that HOST-CP has a lower running time compared to DVRL.}
\label{tab:acctime}
\begin{tabular}{|l|l|l|l|l|l|l|l|}
\hline
\multicolumn{2}{|c|}{\textbf{\begin{tabular}[c]{@{}c@{}}Accuracy\\ (5\%)\end{tabular}}} & \multicolumn{2}{c|}{\textbf{\begin{tabular}[c]{@{}c@{}}Accuracy\\ (20\%)\end{tabular}}} & \multicolumn{2}{c|}{\textbf{\begin{tabular}[c]{@{}c@{}}Epochs (HOST-CP)/\\ Iterations (DVRL)\end{tabular}}} & \multicolumn{2}{c|}{\textbf{\begin{tabular}[c]{@{}c@{}}Time\\ (mins)\end{tabular}}} \\ \hline
\begin{tabular}[c]{@{}l@{}}HOST\\ -CP\end{tabular}                & DVRL                & \begin{tabular}[c]{@{}l@{}}HOST-\\ -CP\end{tabular}                & DVRL               & \begin{tabular}[c]{@{}l@{}}HOST-\\ -CP\end{tabular}                         & DVRL                         & \begin{tabular}[c]{@{}l@{}}HOST-\\ -CP\end{tabular}              & DVRL             \\ \hline
62.46                                                             & 57.8                & 71.32                                                              & 66.4               & 1                                                                           & 25                           & 1.1                                                              & 3.40                 \\ \hline
63.34                                                             & 57.2                & 73.82                                                              & 66.8               & 2                                                                           & 50                           & 2.01                                                             & 5.31                 \\ \hline
65.14                                                             & 57.4                & 74.89                                                              & 66.5               & 3                                                                           & 100                          & 3.05                                                             & 6.50                 \\ \hline
65.71                                                             & 57.6                & 75.0                                                               & 66.6               & 4                                                                           & 150                          & 4.08                                                             & 10.57                 \\ \hline
65.72                                                             & 57.4                & 75.0                                                               & 66.4               & 5                                                                           & 200                          & 5.12                                                             & 14.26                 \\ \hline
\end{tabular}
\end{table}

We analysed the computational complexity of the proposed method on a 64-bit machine with one Quadro-P5000 GPU. For this experiment, we generated synthetic datasets with varying number of training datapoints (500,2000,4000,6000,10000). Keeping the network architecture fixed at a 3 layer network, we varied the size of the training data, on which the subset is meant to be computed. We also varied the size of the test datapoints for which explanations are sought to be found from the training datapoints. We report the time taken to perform the joint subset-training over an epoch, using the proposed method, which aims at finding the best set of explanations from the training set for a given set of test data. Figure \ref{fig:timec} shows a linear pattern in time (in minutes) consumed by the proposed method across the varying size of datapoints.

We have earlier observed that the proposed method is consistently performing better than all the baselines. Since DVRL is the closest to our approach, we also record the time taken by DVRL and the proposed method on the synthetic dataset. We increase the number of iterations for DVRL to trace if the accuracy obtained by the proposed method is achievable by DVRL under a certain subset cardinality (5\%, 20\%). We can observe that DVRL saturates at a certain accuracy value even after increasing the number of iterations to a much higher value. Thus, our method, HOST-CP besides attaining a higher accuracy, also takes considerably lower running time than that of DVRL.




\section{Conclusion}

In this paper, we propose a technique for finding high-value subsets of training datapoints essential for explaining the test set instances. We design a learning convex framework for subset selection, which is then used to optimise the subset with respect to a differentiable value function. The value function is essentially the performance metric which helps to evaluate the trained models. We compare our method against the state-of-the-art baselines , influence functions \cite{koh2017understanding}, data shapley \cite{ghorbani2019data}, TracIn-CP \cite{pruthi2020estimating} and DVRL \cite{yoon2020data} across different datasets in a range of applications like addition or removal of data, detecting mislabelled examples. We also analyse the quality of obtained subsets from DVRL and the proposed method. Lastly, we show that the proposed method scales linearly with the dataset size and takes a lower running time than DVRL which is the closest method to our approach methodologically. Thus, we are able to show that our method, HOST-CP outperforms the baselines consistently in all the applications used for the experiments, thus proving its efficiency in terms of providing high-value subsets.
\label{submission}

\bibliographystyle{splncs04}
\bibliography{ref}

\begin{thebibliography}{10}
\providecommand{\url}[1]{\texttt{#1}}
\providecommand{\urlprefix}{URL }
\providecommand{\doi}[1]{https://doi.org/#1}

\bibitem{agrawal2019differentiable}
Agrawal, A., Amos, B., Barratt, S., Boyd, S., Diamond, S., Kolter, Z.:
  Differentiable convex optimization layers. arXiv preprint arXiv:1910.12430
  (2019)

\bibitem{buchbinder2014submodular}
Buchbinder, N., Feldman, M., Naor, J., Schwartz, R.: Submodular maximization
  with cardinality constraints. In: Proceedings of the twenty-fifth annual
  ACM-SIAM symposium on Discrete algorithms. pp. 1433--1452. SIAM (2014)

\bibitem{choromanska2019beyond}
Choromanska, A., Cowen, B., Kumaravel, S., Luss, R., Rigotti, M., Rish, I.,
  Diachille, P., Gurev, V., Kingsbury, B., Tejwani, R., et~al.: Beyond
  backprop: Online alternating minimization with auxiliary variables. In:
  International Conference on Machine Learning. pp. 1193--1202. PMLR (2019)

\bibitem{cook1982residuals}
Cook, R.D., Weisberg, S.: Residuals and influence in regression. New York:
  Chapman and Hall (1982)

\bibitem{das2020multi}
Das, S., Mandal, S., Bhoyar, A., Bharde, M., Ganguly, N., Bhattacharya, S.,
  Bhattacharya, S.: Multi-criteria online frame-subset selection for autonomous
  vehicle videos. Pattern Recognition Letters  (2020)

\bibitem{elhamifar2017online}
Elhamifar, E., Kaluza, M.C.D.P.: Online summarization via submodular and convex
  optimization. In: CVPR. pp. 1818--1826 (2017)

\bibitem{elhamifar2015dissimilarity}
Elhamifar, E., Sapiro, G., Sastry, S.S.: Dissimilarity-based sparse subset
  selection. IEEE transactions on pattern analysis and machine intelligence
  \textbf{38}(11),  2182--2197 (2015)

\bibitem{frenay2013classification}
Fr{\'e}nay, B., Verleysen, M.: Classification in the presence of label noise: a
  survey. IEEE transactions on neural networks and learning systems
  \textbf{25}(5),  845--869 (2013)

\bibitem{ghorbani2020distributional}
Ghorbani, A., Kim, M., Zou, J.: A distributional framework for data valuation.
  In: International Conference on Machine Learning. pp. 3535--3544. PMLR (2020)

\bibitem{ghorbani2019data}
Ghorbani, A., Zou, J.: Data shapley: Equitable valuation of data for machine
  learning. In: International Conference on Machine Learning. pp. 2242--2251.
  PMLR (2019)

\bibitem{ghorbani2020neuron}
Ghorbani, A., Zou, J.: Neuron shapley: Discovering the responsible neurons.
  arXiv preprint arXiv:2002.09815  (2020)

\bibitem{hara2019data}
Hara, S., Nitanda, A., Maehara, T.: Data cleansing for models trained with sgd.
  arXiv preprint arXiv:1906.08473  (2019)

\bibitem{koh2017understanding}
Koh, P.W., Liang, P.: Understanding black-box predictions via influence
  functions. In: International Conference on Machine Learning. pp. 1885--1894.
  PMLR (2017)

\bibitem{lundberg2017unified}
Lundberg, S., Lee, S.I.: A unified approach to interpreting model predictions.
  arXiv preprint arXiv:1705.07874  (2017)

\bibitem{lundberg2018consistent}
Lundberg, S.M., Erion, G.G., Lee, S.I.: Consistent individualized feature
  attribution for tree ensembles. arXiv preprint arXiv:1802.03888  (2018)

\bibitem{pruthi2020estimating}
Pruthi, G., Liu, F., Kale, S., Sundararajan, M.: Estimating training data
  influence by tracing gradient descent. Advances in Neural Information
  Processing Systems  \textbf{33} (2020)

\bibitem{steinhardt2017certified}
Steinhardt, J., Koh, P.W., Liang, P.: Certified defenses for data poisoning
  attacks. arXiv preprint arXiv:1706.03691  (2017)

\bibitem{wu2020deltagrad}
Wu, Y., Dobriban, E., Davidson, S.: Deltagrad: Rapid retraining of machine
  learning models. In: International Conference on Machine Learning. pp.
  10355--10366. PMLR (2020)

\bibitem{yoon2020data}
Yoon, J., Arik, S., Pfister, T.: Data valuation using reinforcement learning.
  In: International Conference on Machine Learning. pp. 10842--10851. PMLR
  (2020)

\end{thebibliography}




\end{document}